\documentclass[sigconf]{acmart}

\copyrightyear{2023}
\acmYear{2023}
\setcopyright{acmlicensed}\acmConference[GECCO '23]{Genetic and Evolutionary Computation Conference}{July 15-19, 2023}{Lisbon, Portugal}
\acmBooktitle{Genetic and Evolutionary Computation Conference (GECCO '23), July 15-19, 2023, Lisbon, Portugal}
\acmPrice{15.00}
\acmDOI{10.1145/3583131.3590518}
\acmISBN{979-8-4007-0119-1/23/07}

\usepackage{subcaption}
\usepackage{algorithm}
\usepackage{algorithmic}
\usepackage{bm}

\newcommand{\secref}[1]{Section\ \ref{#1}}
\newcommand{\figref}[1]{Fig.\ \ref{#1}}
\newcommand{\tabref}[1]{Table\ \ref{#1}}
\newcommand{\algoref}[1]{Algorithm\ \ref{#1}}
\newcommand{\eqaref}[1]{Eq.\ (\ref{#1})}

\begin{document}

\title{Natural Evolution Strategy for\\Mixed-Integer Black-Box Optimization}

\author{Koki Ikeda}
\email{ikeda.k@ic.c.titech.ac.jp}
\affiliation{%
  \institution{Tokyo Institute of Technology}
  \city{Yokohama}
  \state{Kanagawa}
  \country{Japan}
}

\author{Isao Ono}
\email{isao@c.titech.ac.jp}
\affiliation{%
  \institution{Tokyo Institute of Technology}
  \city{Yokohama}
  \state{Kanagawa}
  \country{Japan}
}

\begin{abstract}
  \sloppy
  This paper proposes a natural evolution strategy (NES) for mixed-integer black-box optimization (MI-BBO) that appears in real-world problems such as hyperparameter optimization of machine learning and materials design.
  This problem is difficult to optimize because plateaus where the values do not change appear when the integer variables are relaxed to the continuous ones.
  CMA-ES w. Margin that addresses the plateaus reportedly showed good performance on MI-BBO benchmark problems.
  However, it has been observed that the search performance of CMA-ES w. Margin deteriorates when continuous variables contribute more to the objective function value than integer ones.
  In order to address the problem of CMA-ES w. Margin, we propose Distance-weighted eXponential Natural Evolution Strategy taking account of Implicit Constraint and Integer (DX-NES-ICI).
  We compare the search performance of DX-NES-ICI with that of CMA-ES w. Margin through numerical experiments.
  As a result, DX-NES-ICI was up to 3.7 times better than CMA-ES w. Margin in terms of a rate of finding the optimal solutions on benchmark problems where continuous variables contribute more to the objective function value than integer ones.
  DX-NES-ICI also outperformed CMA-ES w. Margin on problems where CMA-ES w. Margin originally showed good performance.
\end{abstract}



\maketitle

\section{introduction}\label{sec:introduction}
The mixed-integer black-box optimization (MI-BBO) problem is a problem of optimizing a black-box function whose decision variables consist of continuous and integer ones.
In MI-BBO, the objective function is not explicitly given, which makes the problem difficult to solve because the gradient of the objective function cannot be analytically calculated and assumptions such as convexity cannot be made.
MI-BBO problems often appear in real-world problems such as hyperparameter optimization of machine learning\cite{hazan18, hutter19}, materials design\cite{iyer20, zhang20}, topology optimization\cite{fujii18, yang98}, placement optimization for carbon dioxide capture and storage\cite{miyagi18}, and lens design \cite{ono00}.

MI-BBO methods have not been developed as aggressively as continuous or discrete BBO methods, despite the high demand for efficient ones.
Continuous BBO methods can be applied to MI-BBO problems by relaxing the integer variables to the continuous ones.
However, plateaus where the values do not change appear by the continuous relaxation, making optimization by continuous BBO methods difficult\cite{cmaeswm}.
For example, CMA-ES\cite{hansen03, hansen96} repeats the operations of generating solutions according to a multivariate Gaussian distribution (MGD) and updating the MGD so that better solutions are more likely to be generated to search for the optimum.
If the MGD comes to lie within a plateau in an integer variable dimension, the integer variables of all the solutions generated according to the MGD are equal to each other. In this situation, the direction of the improvement is not be found in the integer variable dimension. As a result, the MGD will be trapped in the plateau in the integer variable dimension.
Several algorithms, including an extended evolution strategy\cite{li13}, a surrogate model-based method\cite{bliek21}, and CMA-ES w. Margin\cite{cmaeswm} have been proposed for MI-BBO problems.

CMA-ES w. Margin is an MI-BBO method, based on CMA-ES, that addresses the above problem of CMA-ES when applied to MI-BBO \cite{cmaeswm}.
Even when the MGD is trapped in a plateau, CMA-ES w. Margin moves its MGD in the direction of improvement in an integer variable dimension by obtaining different integer values for calculating objective function values from the neighbor plateaus.
In order to obtain the different values, CMA-ES w. Margin uses the affine transformation of solutions and the mean correction.
However, CMA-ES w. Margin has a problem in that the search performance deteriorates for functions where continuous variables contribute more to the objective function values than integer ones.
When CMA-ES w. Margin is applied to those functions, the size of MGD keeps decreasing because the search proceeds further for the continuous variable dimensions than for the integer variable dimensions in the early stage of the search.
As a result, the size of the MGD would become much smaller than that of a plateau.
In this situation, even if CMA-ES w. Margin obtains different integer values for calculating objective function values from the neighbor plateaus, it cannot sufficiently update the MGD so as to move it to the next plateau.
MI-BBO methods are expected to perform well regardless of whether integer or continuous variables contribute more to the objective function value than the others because it cannot be known in advance in MI-BBO.

In this paper, in order to address the above problem of CMA-ES w. Margin, we propose Distance-weighted eXponential Natural Evolution Strategy taking account of Implicit Constraint and Integer (DX-NES-ICI) based on DX-NES-IC \cite{dxnesic}.
DX-NES-IC is one of the promising NESs and has been reported to outperform CMA-ES on continuous BBO benchmark problems \cite{dxnesic}.
In order to address the problem of CMA-ES w. Margin, DX-NES-ICI always places the MGD on the boundary of two plateaus when the size of the MGD becomes smaller than that of a plateau.
Then, DX-NES-ICI moves the MGD to the next boundary if the overlap between the MGD and one of the two plateaus becomes small in order to prevent the MGD from being trapped in a plateau.
We compare the search performance of DX-NES-ICI with that of CMA-ES w. Margin using benchmark problems where continuous variables contribute more to the objective function value than integer ones and the other benchmark problems where CMA-ES w. Margin originally showed good performance.

The rest of the paper is as follows.
Section 2 defines the MI-BBO problem handled in this paper.
Section 3 explains CMA-ES w. Margin and points out its problem.
Section 4 proposes a NES to address the problem of CMA-ES w. Margin.
In Section 5, performance comparison experiments are conducted using some benchmark problems to demonstrate the effectiveness of DX-NES-ICI.
Section 6 is discussion, and Section 7 is conclusion.

\section{Mixed-Integer Black-Box Optimization Problem}\label{sec:mibboProblem}\sloppy
The mixed-integer black-box optimization (MI-BBO) problem is a mixed-integer optimization problem where the objective function is black-box.
The problem is defined by
\begin{eqnarray}
  \underset{\bar{\textbf{x}}}{\text{minimize}}\  f(\bar{\textbf{x}}),
\end{eqnarray}
where \(f\) is an \(N\)-dimensional black-box objective function,
\(\bar{\textbf{x}}\) is given by \(\bar{\textbf{x}}:=[\bar{\textbf{x}}_\text{co}^\top,\bar{\textbf{x}}_\text{int}^\top]^\top\),
\(\bar{\textbf{x}}_\text{co}=[\bar{x}_1,\ldots,\bar{x}_k]^\top\) is an \(N_\text{co}\)-dimensional continuous variable vector,
\(\bar{\textbf{x}}_\text{int}=[\bar{x}_{k+1},\ldots,\bar{x}_N]^\top\) is an \(N_\text{int}\)-dimensional integer variable one, \(k=N_\text{co}\), and \(N=N_\text{co}+N_\text{int}\).
In this paper, we assume that \(N_\text{co}=N_\text{int}=N/2\) and the domain \(\bm{z}_j\) of the \(j\)-th variable \([\bar{\textbf{x}}]_j\ (N_\textrm{co}+1\le j \le N)\) is given by
\begin{eqnarray}
  \bm{z}_j=\{z_{j,1},z_{j,2},\ldots,z_{j,K_j}\mid z_{j,1}\le z_{j,2}\le \cdots \le z_{j,K_j}\},
\end{eqnarray}
where \(K_j\) is the number of elements.

\section{CMA-ES w. Margin and its problem}\label{sec:cmaeswm}
\subsection{CMA-ES w. Margin}\label{sec:cmaeswm:method}
CMA-ES w. Margin \cite{cmaeswm} is a promising MI-BBO method based on CMA-ES \cite{hansen03, hansen96} that is one of the state-of-the-art continuous BBO methods.
CMA-ES w. Margin searches in the continuous space based on CMA-ES and transforms a solution \(\textbf{x}\) to \(\bar{\textbf{x}} \ (=\textrm{Encoding}_f (\textbf{x}))\) when \(\textbf{x}\) is evaluated, where \(\textrm{Encoding}_f\) is a transformation function.
If \(N_\textrm{co}+1\le j \le N\), \(\text{Encoding}_f\) is defined by
\begin{eqnarray}
  \textrm{Encoding}_f ([\textbf{x}]_j)=\left\{
  \begin{array}{ll}
    z_{j,1}   & \textrm{if } [\textbf{x}]_j \le \ell_{j,1|2}                    \\
    z_{j,k}   & \textrm{if } \ell_{j,k-1|k} < [\textbf{x}]_j \le \ell_{j,k|k+1} \\
    z_{j,K_j} & \textrm{if } \ell_{j,K_j-1|K_j} < [\textbf{x}]_j
  \end{array}\right.,
\end{eqnarray}
where \([\textbf{x}]_j\) is the \(j\)-th element of \(\textbf{x}\), \(1<k<K_j\), and \(\ell_{j,k|k+1}\in\bm{\ell}_j\) are thresholds given by \(\ell_{j,k|k+1}:=(z_{j,k}+z_{j,k+1})/2\).
Note that the regions \((-\infty, \ell_{j,1|2}]\), \((\ell_{j,k-1|k}, \ell_{j,k|k+1}]\), and \((\ell_{j,K_j-1|K_j}, \infty)\) are plateaus where the values of the integer variables do not change and that the thresholds are the boundaries of the plateaus, and the values of the integer variables change at these points.
On the other hand, if \(1\le j \le N_\textrm{co}\), \(\text{Encoding}_f\) is defined by
\begin{eqnarray}
  \textrm{Encoding}_f ([\textbf{x}]_j)=[\textbf{x}]_j.
\end{eqnarray}

Although CMA-ES can be directly applied to MI-BBO problems by using \(\text{Encoding}_f\), the search performance deteriorates due to the plateaus.
CMA-ES repeats the operations of generating solutions according to a multivariate Gaussian distribution (MGD) \(\mathcal{N}(\textbf{m}, \sigma^2\textbf{C})\) and updating the mean \(\textbf{m}\in\mathbb{R}^N\), the step size \(\sigma\in\mathbb{R}\), and the covariance matrix \(\textbf{C}\in\mathbb{R}^{N\times N}\) so that better solutions are more likely to be generated to search for the optimum.
If the MGD comes to lie within a plateau in an integer variable dimension, the integer variables of all the solutions generated according to the MGD are equal to each other. In this situation, the direction of the improvement is not be found in the integer variable dimension. As a result, the MGD will be trapped in the plateau in the integer variable dimension.

\begin{figure}[tb]
  \centering
  \begin{tabular}{cc}
    \begin{minipage}[t]{0.46\hsize}
      \centering
      \includegraphics[width=39mm]{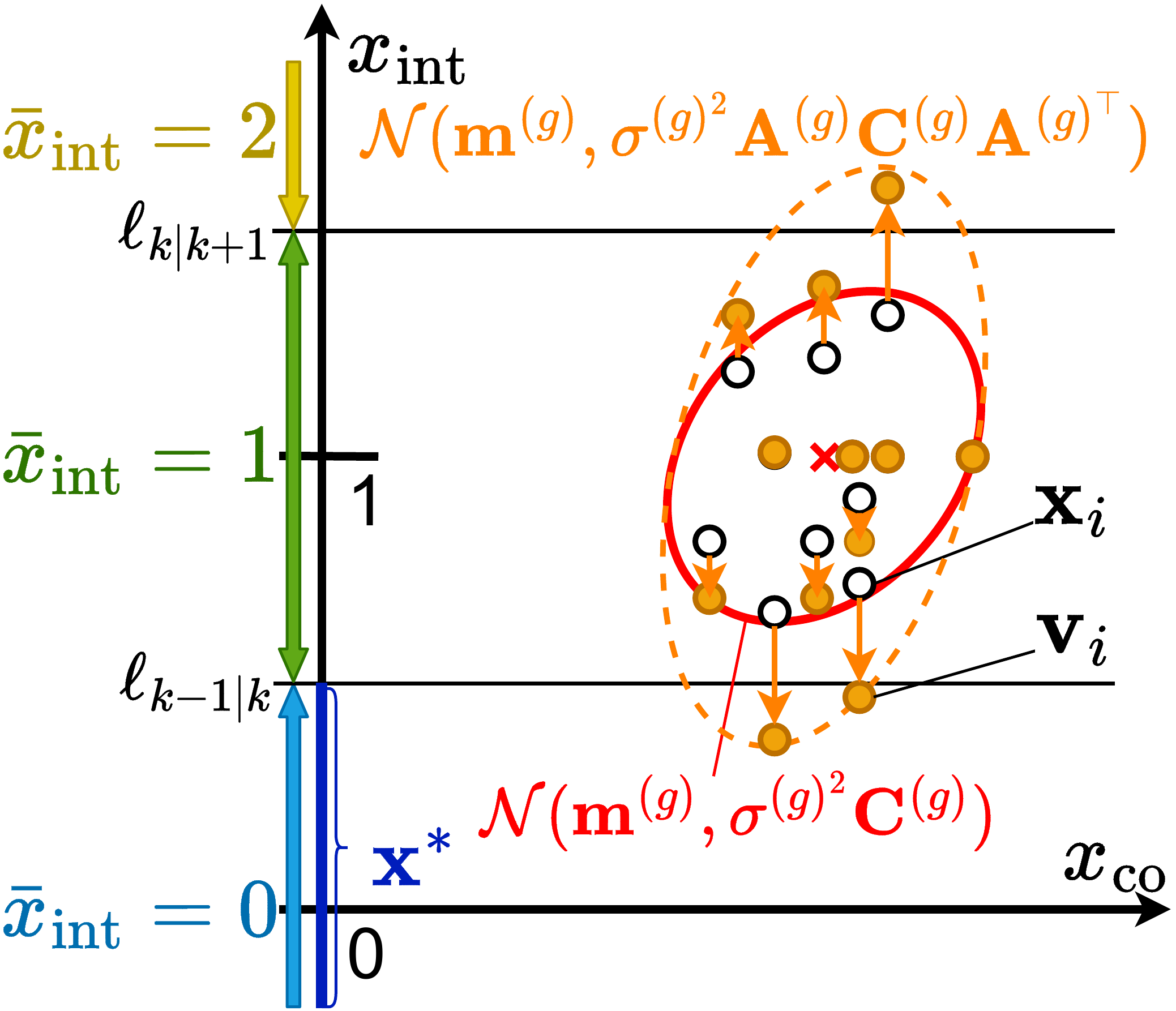}
      \subcaption{\(\{\textbf{x}_i\}_{i = 1}^\lambda\) are generated according to MGD in $g$-th generation, and each \(\textbf{x}_i\) is evaluated by the evaluation value of \(\textbf{v}_i\).}
      \label{fig:cmaeswmAlgorithm:evaluation}
    \end{minipage}
     &
    \begin{minipage}[t]{0.46\hsize}
      \centering
      \includegraphics[width=39mm]{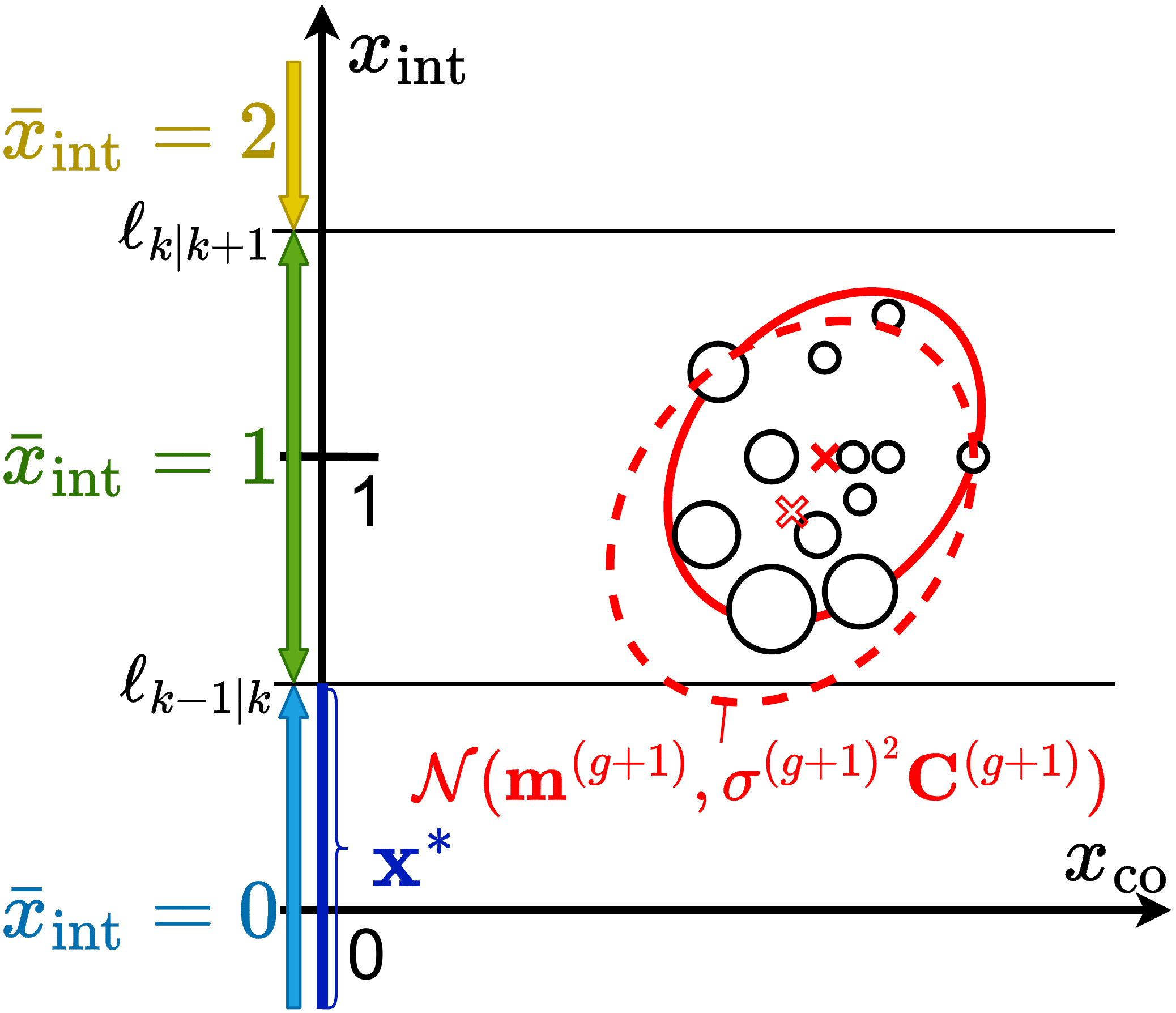}
      \subcaption{\(\{\textbf{x}_i\}_{i = 1}^\lambda\) are weighted according to their evaluation values, and the parameters of MGD are updated based on the weights of \(\{\textbf{x}_i\}_{i = 1}^\lambda\) to obtain the next MGD.}
      \label{fig:cmaeswmAlgorithm:update}
    \end{minipage}
  \end{tabular}
  \caption{Behavior of CMA-ES w. Margin.
            The vertical axis, the horizontal axis, \(\textbf{x}^*\), the red ellipse, the white circles, and the orange circles represent
            the dimension of an integer variable, the dimension of a continuous variable, the optimal region, MGD \(\mathcal{N}(\textbf{m}, \sigma^2\textbf{C})\), the generated solutions \(\{\textbf{x}_i\}_{i=1}^\lambda\), and the solutions \(\{\textbf{v}_i\}_{i=1}^\lambda\) obtained by applying an affine transformation to \(\{\textbf{x}_i\}_{i=1}^\lambda\), respectively.
  }
  \label{fig:cmaeswmAlgorithm}
\end{figure}

\begin{figure}[tb]
  \centering
  \begin{tabular}{cc}
    \begin{minipage}[t]{0.46\hsize}
      \centering
      \includegraphics[width=30mm]{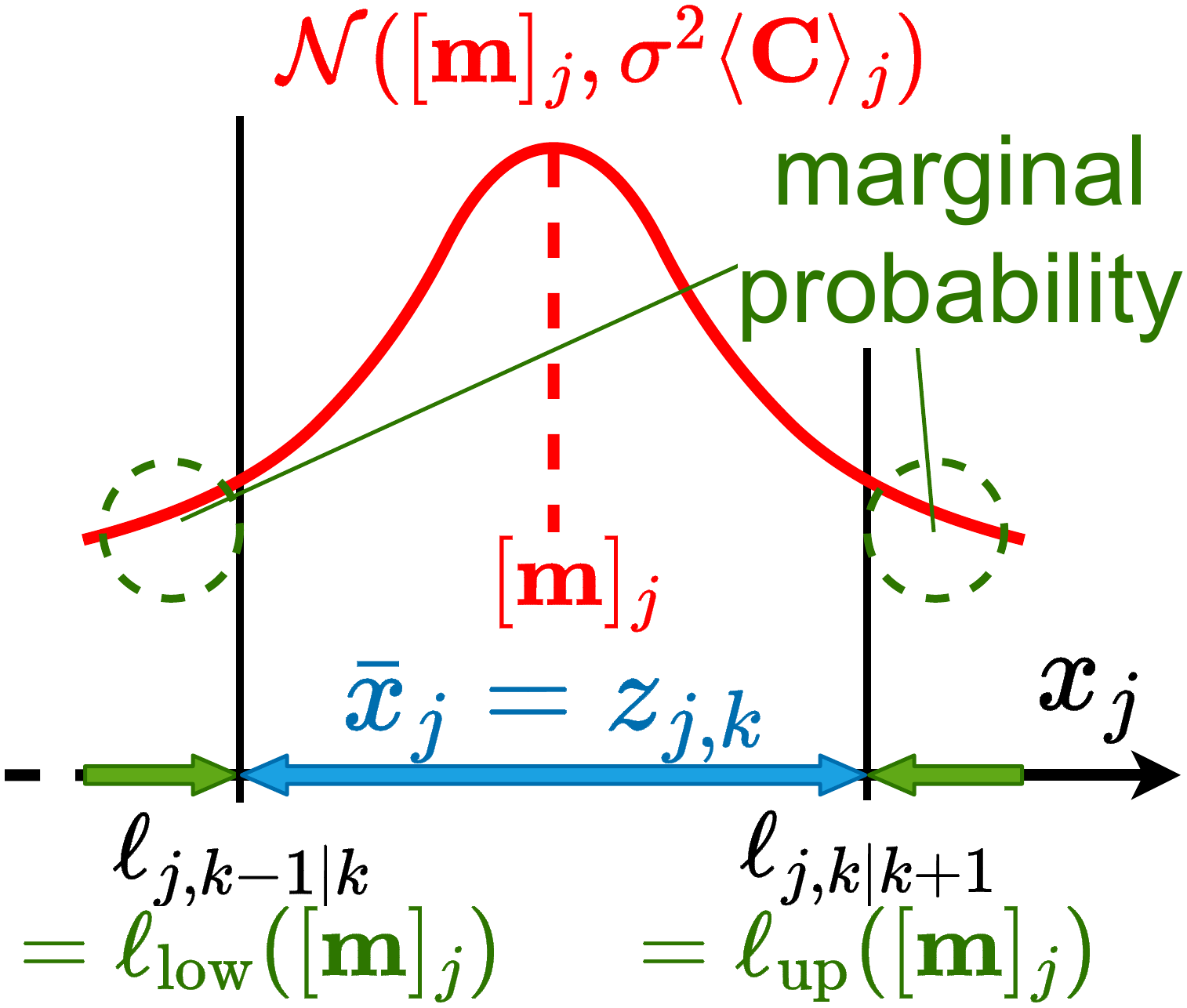}
      \subcaption{Marginal probabilities \(\text{Pr}(\ell_\textrm{up}\left([\textbf{m}]_j\right) < x_j)\) (the right green dotted circle) and \(\text{Pr}(x_j \le \ell_\textrm{low}\left([\textbf{m}]_j\right))\) (the left green dotted circle)}
      \label{fig:cmaeswmMargin:twoSide}
    \end{minipage}
     &
    \begin{minipage}[t]{0.46\hsize}
      \centering
      \includegraphics[width=30mm]{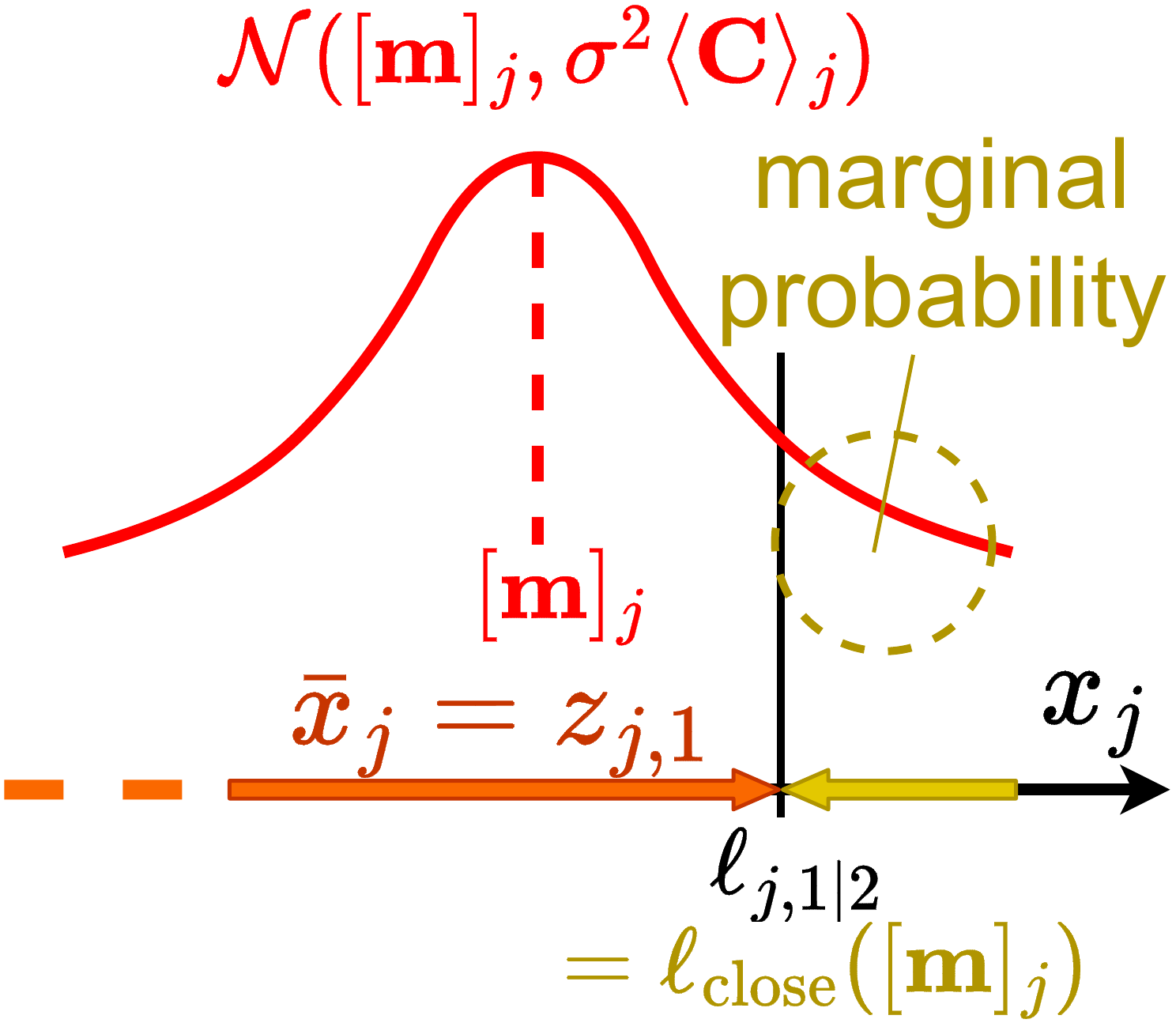}
      \subcaption{A marginal probability \(\text{Pr}(\ell_\textrm{close}\left([\textbf{m}]_j\right) < x_j)\)}
      \label{fig:cmaeswmMargin:oneSide}
    \end{minipage}
  \end{tabular}
  \caption{Example of marginal probabilities in the \(j\)-th dimension.
    The horizontal axis and the red curve represent the dimension of an integer variable and a normal distribution \(\mathcal{N}([\textbf{m}]_j, \sigma^2\langle\textbf{C}\rangle_j)\), respectively.
  }
  \label{fig:cmaeswmMargin}
\end{figure}

In order to prevent the MGD from being trapped in a plateau, as shown in \figref{fig:cmaeswmAlgorithm}, if the MGD is determined to be being trapped in the plateau, CMA-ES w. Margin updates the MGD by using solutions \(\{\textbf{x}_i\}_{i=1}^\lambda\) with objective function values \(\{f(\bar{\textbf{v}}_i)\}_{i=1}^\lambda\), where \(\bar{\textbf{v}}_i=\textrm{Encoding}_f (\textbf{v}_i)\) and \(\textbf{v}_i\) are obtained by applying an affine transformation to \(\textbf{x}_i\), and after updating the MGD, moves the mean of the MGD by the mean correction method.
CMA-ES w. Margin determines that the MGD is being trapped in a plateau in the \(j\)-th dimension if each marginal probability of the MGD, shown in \figref{fig:cmaeswmMargin:twoSide}, is less than \(\alpha/2\), where \(\alpha\) is a user parameter.
The marginal probabilities are an upper tail probability \(\text{Pr}(\ell_\textrm{up}\left([\textbf{m}]_j\right) < x_j)\) and a lower tail probability \(\text{Pr}(x_j \le \ell_\textrm{low}\left([\textbf{m}]_j\right))\) of a normal distribution \(\mathcal{N}([\textbf{m}]_j, \sigma^2\langle\textbf{C}\rangle_j)\) given by
\begin{eqnarray}
  \ell_\textrm{low}\left([\textbf{m}]_j\right) &:=& \max\left\{l \in \bm{\ell}_j\mid l<[\textbf{m}]_j\right\},\\
  \ell_\textrm{up}\left([\textbf{m}]_j\right)  &:=& \min\left\{l \in \bm{\ell}_j\mid [\textbf{m}]_j\le l\right\},
\end{eqnarray}
where \(\langle\cdot\rangle_j\) is the \(j\)-th diagonal element of an argument matrix.
The affine transformation is defined as a matrix that magnifies the MGD so as to ensure that  \(\text{Pr}(\ell_\textrm{up}\left([\textbf{m}]_j\right) < x_j) \ge \alpha/2\) and \(\text{Pr}(x_j \le \ell_\textrm{low}\left([\textbf{m}]_j\right)) \ge \alpha/2\).
The mean correction method moves the mean vector of the MGD so as to increase the smaller marginal probability.

When \([\textbf{m}]_j \le \ell_{j,1|2}\) or \(\ell_{j,K_j-1|K_j}<[\textbf{m}]_j \), a marginal probability of only one side is defined, as shown in \figref{fig:cmaeswmMargin:oneSide}.
The marginal probability is a cumulative probability \(\text{Pr}(\ell_\textrm{close}\left([\textbf{m}]_j\right) < x_j)\) or \(\text{Pr}(x_j \le \ell_\textrm{close}\left([\textbf{m}]_j\right))\), where \(\ell_\textrm{close}\left([\textbf{m}]_j\right)\) is defined as the threshold closest to \([\textbf{m}]_j\).
In this situation, the same affine transformation as in the previous generation is used to move solutions for evaluating them.
The mean correction method moves the mean vector of the MGD so as to increase the marginal probability as follows.
\begin{eqnarray}
  \label{eq:cmaeswm:modifyM}
  [\textbf{m}^{\text{after}}]_j & \leftarrow
  & \ell_\text{close} \left([\textbf{m}^{\text{after}}]_j\right)+\nonumber\\
  & & {\rm sign}\left([\textbf{m}^{\text{after}}]_j - \ell_\text{close}\left([\textbf{m}^{\text{after}}]_j\right)\right) \nonumber \\
  & & \times {\rm CI}_j((\sigma^{\text{after}})^2 \textbf{C}^{\text{after}}, \alpha),\\
  \textrm{CI}_j(\boldsymbol{\Sigma}, \alpha)&:=&\mathcal{N}(0,1)_\text{ppf}(\alpha) \sqrt{\langle \boldsymbol{\Sigma} \rangle_j},
\end{eqnarray}
where \(\textbf{m}^{\text{after}}\), \(\sigma^{\text{after}}\), and \(\textbf{C}^{\text{after}}\) are the mean vector, the step size, the covariance matrix of the MGD after updated by CMA-ES, respectively, and
\(\mathcal{N}(0,1)_\text{ppf}(\alpha)\) is a percentage point at which the upper tail probability of the standard normal distribution is \(\alpha\).
Note that \(\left[[\textbf{m}^{\text{after}}]_j - \textrm{CI}_j(\boldsymbol{\Sigma}, \alpha), [\textbf{m}^{\text{after}}]_j + \textrm{CI}_j(\boldsymbol{\Sigma}, \alpha)\right]\) is a confidence interval with level \(\alpha\).

\subsection{Problem of CMA-ES w. Margin}
\begin{figure}[tb]
  \centering
  \begin{tabular}{cc}
    \begin{minipage}[t]{0.46\hsize}
      \centering
      \includegraphics[width=39mm]{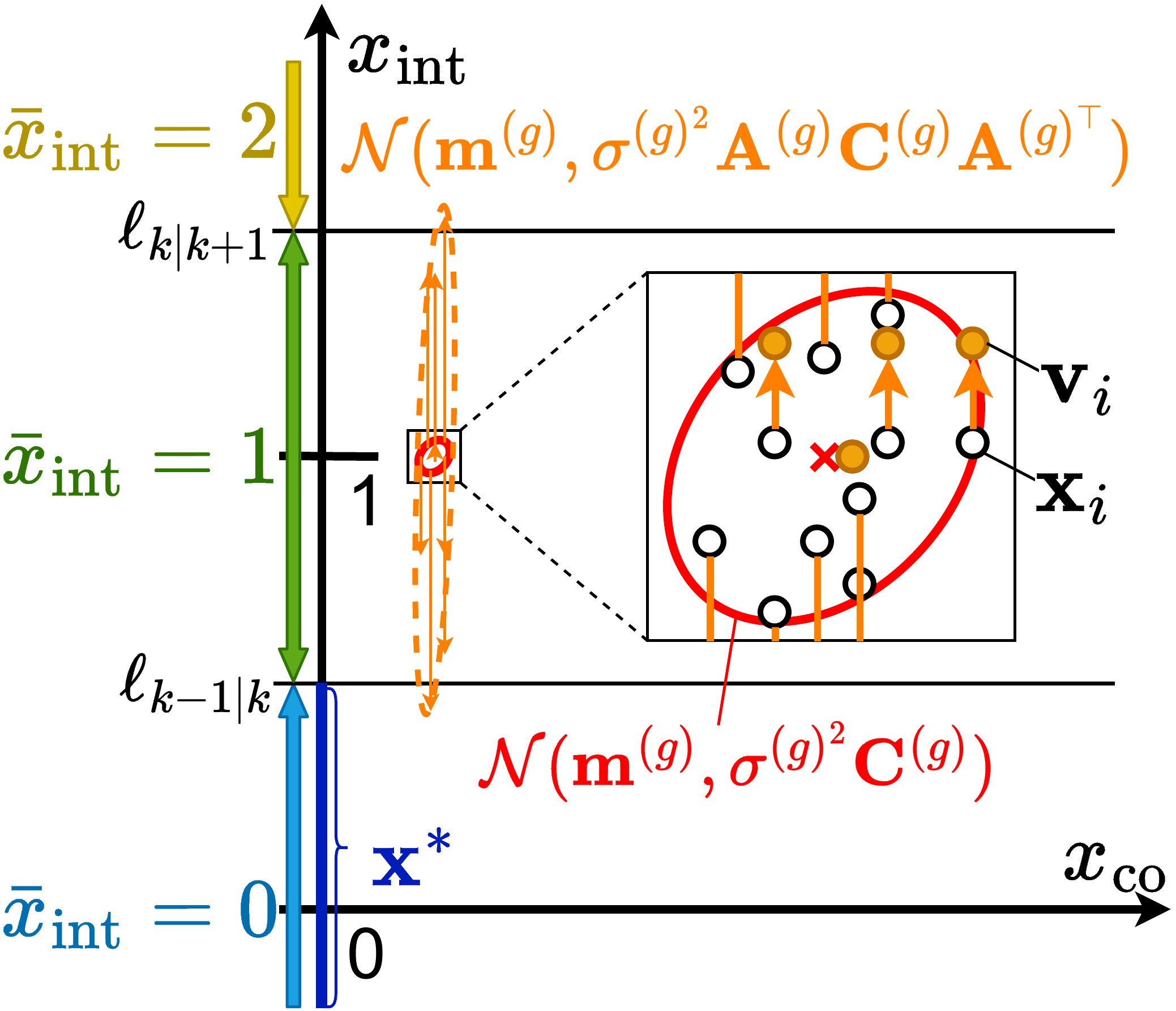}
      \subcaption{The same operation as \figref{fig:cmaeswmAlgorithm:evaluation}.}
      \label{fig:cmaeswmStagnation:evaluation}
    \end{minipage}
     &
    \begin{minipage}[t]{0.46\hsize}
      \centering
      \includegraphics[width=39mm]{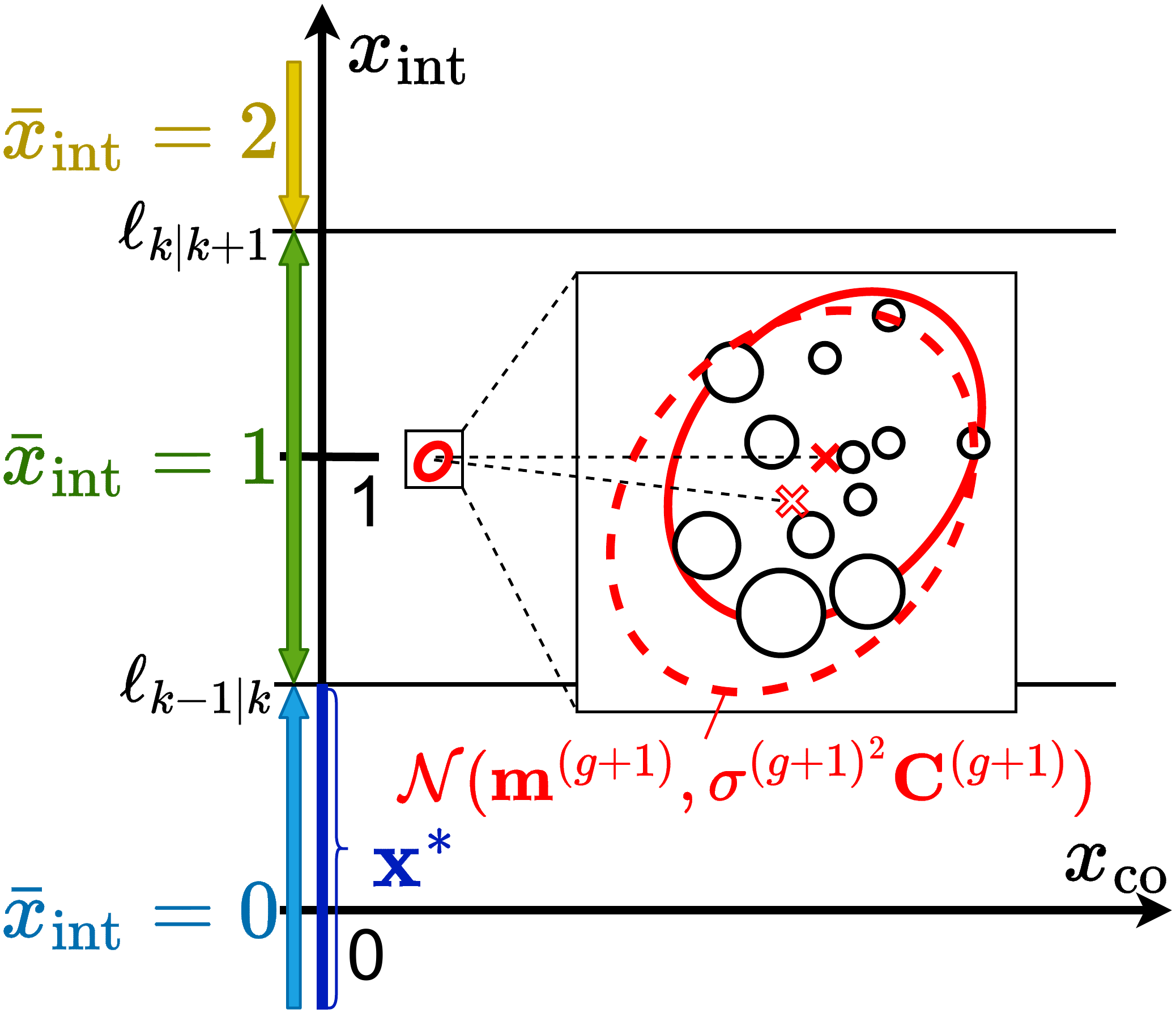}
      \subcaption{The same operation as \figref{fig:cmaeswmAlgorithm:update}.}
      \label{fig:cmaeswmStagnation:update}
    \end{minipage}
  \end{tabular}
  \caption{How the CMA-ES w. Margin stagnates in the plateau for functions where continuous variables contribute more to the objective function value than integer ones.}
  \label{fig:cmaeswmStagnation}
\end{figure}

CMA-ES w. Margin has a problem in that the search performance deteriorates for functions where continuous variables contribute more to the objective function values than integer ones.
When CMA-ES w. Margin is applied to those functions, the size of MGD keeps decreasing because the search proceeds further for the continuous variable dimensions than for the integer variable dimensions in the early stage of the search.
As a result, the size of the MGD would become much smaller than that of a plateau, as shown in \figref{fig:cmaeswmStagnation}.
In this situation, even if CMA-ES w. Margin obtains different integer values for calculating objective function values from the neighbor plateaus as shown in \figref{fig:cmaeswmStagnation:evaluation}, it cannot sufficiently update the MGD so as to move it to the next plateau as shown in \figref{fig:cmaeswmStagnation:update}.
As a result, it will take a long time for the MGD to escape from the plateau.

MI-BBO methods are expected to perform well regardless of whether integer or continuous variables contribute more to the objective function value than the others because it cannot be known in advance in MI-BBO problems.

\section{DX-NES-ICI}\label{proposed}
\subsection{Basic ideas}
In this paper, in order to address the above problem of CMA-ES w. Margin, we propose Distance-weighted eXponential Natural Evolution Strategy taking account of Implicit Constraint and Integer (DX-NES-ICI) based on DX-NES-IC \cite{dxnesic}.
DX-NES-IC is one of the promising NESs and has been reported to outperform CMA-ES on continuous BBO benchmark problems \cite{dxnesic}.
DX-NES-ICI employs DX-NES-IC to search in the continuous space and the transformation function \(\text{Encoding}_f\), as CMA-ES w. Margin does, to transform a continuous solution vector \(\textbf{x}\) to a mixed-integer solution vector \(\bar{\textbf{x}}\) for evaluation.

\begin{figure}[tb]
    \centering
    \begin{tabular}{cc}
        \begin{minipage}[t]{0.46\hsize}
            \centering
            \includegraphics[width=39mm]{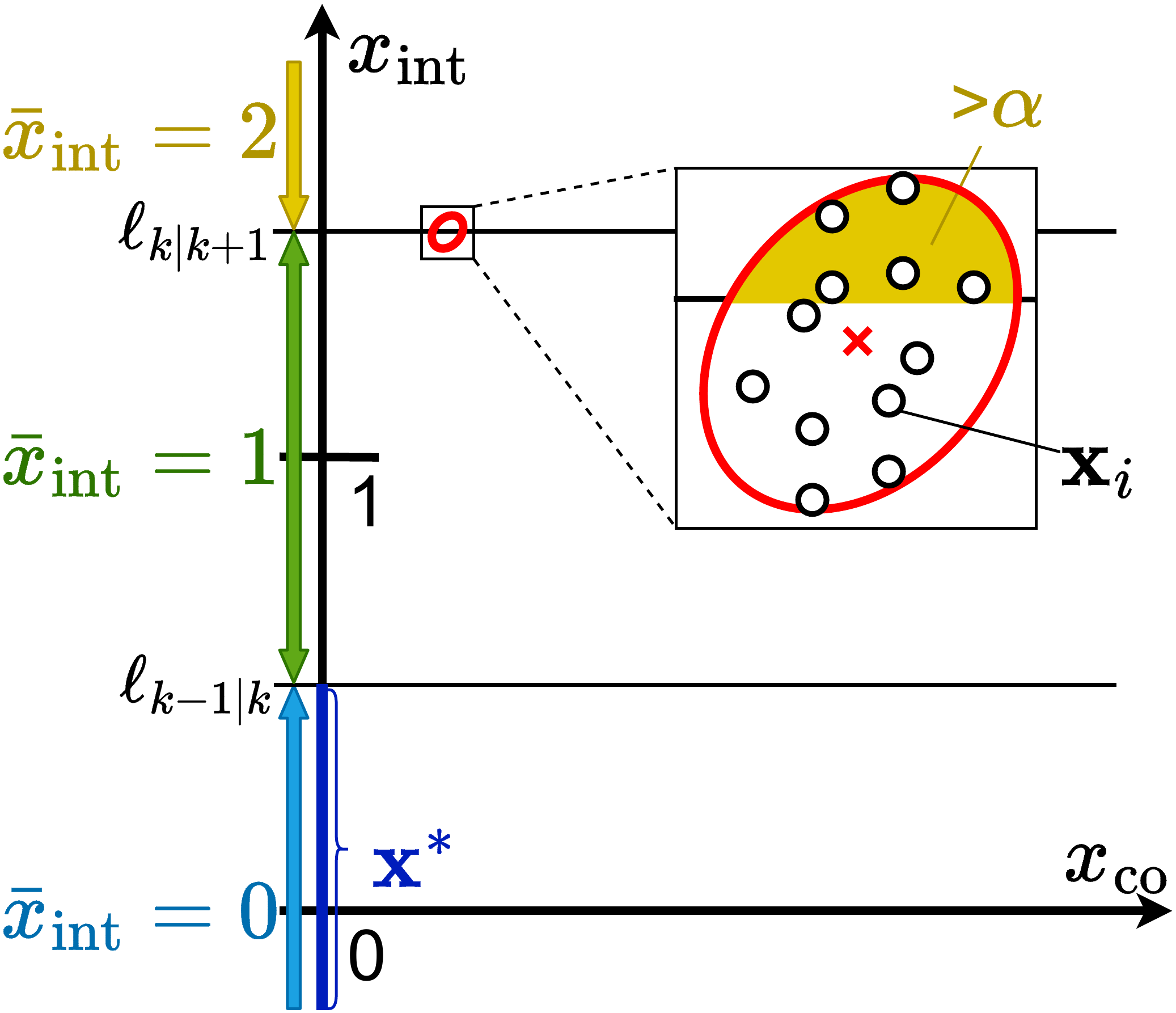}
            \subcaption{DX-NES-ICI computes the evaluation values of \(\{\textbf{x}_i\}_{i=1}^\lambda\) generated according to MGD.}
            \label{fig:mywayAlgorithm:evaluation1}
        \end{minipage}
         &
        \begin{minipage}[t]{0.46\hsize}
            \centering
            \includegraphics[width=39mm]{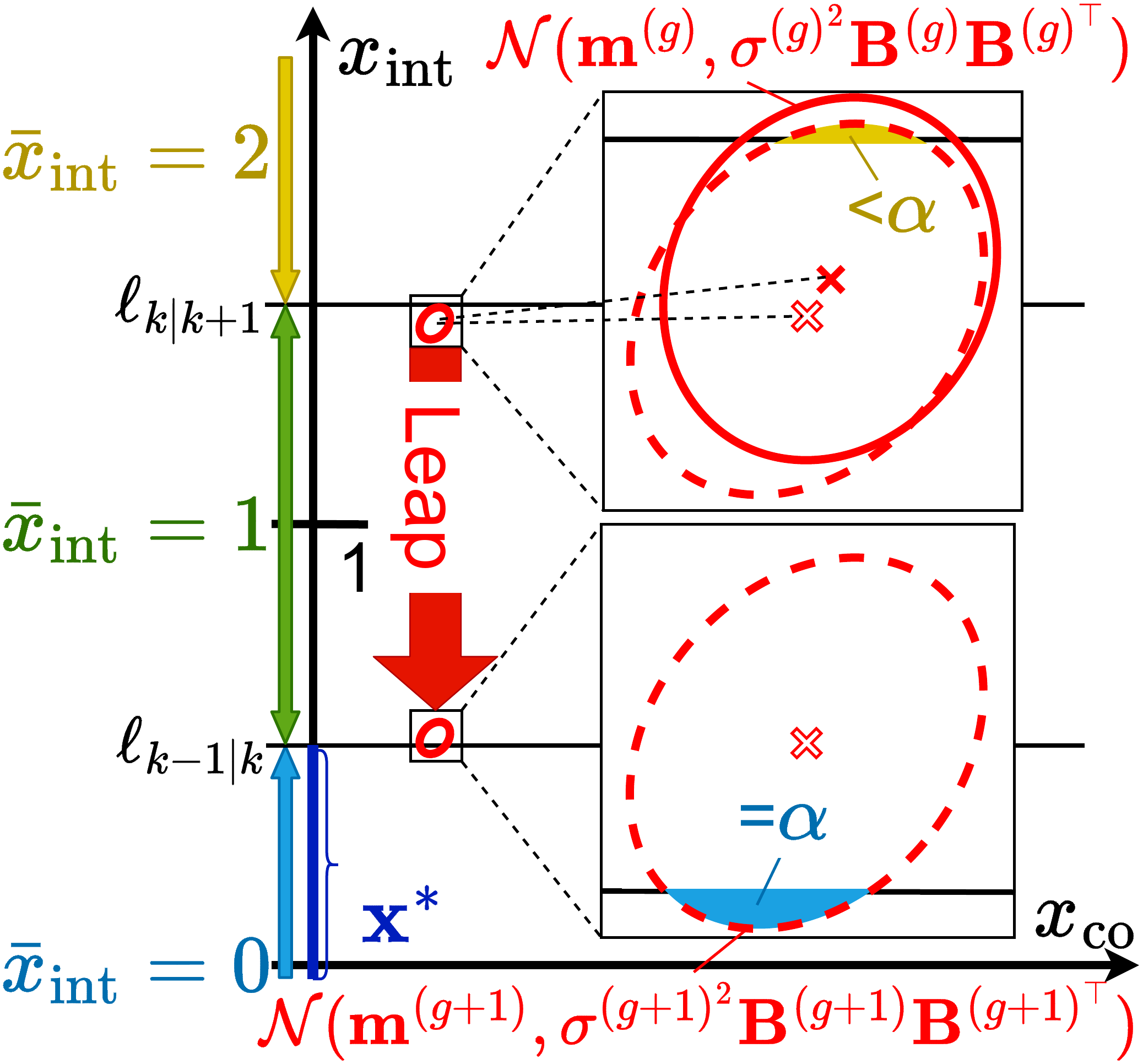}
            \subcaption{As a result of updating the parameters of MGD in $g$-th generation, a marginal probability becomes less than \(\alpha\), and the mean vector leaps to obtain the next MGD.}
            \label{fig:mywayAlgorithm:update1}
        \end{minipage}
    \end{tabular}
    \caption{Behavior of DX-NES-ICI when applied to functions where continuous variables contribute more to the objective function value than integer ones.
            The red ellipse, the white circles, and the red arrow represent
            MGD \(\mathcal{N}(\textbf{m}, \sigma^2\textbf{B}\textbf{B}^\top)\), the generated solutions \(\{\textbf{x}_i\}_{i=1}^\lambda\), and the operation of leaping the mean vector, respectively.}
    \label{fig:mywayAlgorithm}
\end{figure}

In order to address the problem of CMA-ES w. Margin, DX-NES-ICI always places the MGD on the boundary of two plateaus when the size of the MGD is smaller than that of a plateau, as shown in \figref{fig:mywayAlgorithm:evaluation1}.
Then, if the overlap between the MGD and one of the two plateaus becomes small, i.e., the marginal probability is less than \(\alpha\), DX-NES-ICI moves the MGD to the next boundary in order to prevent the MGD from being trapped in the plateau as shown in \figref{fig:mywayAlgorithm:update1}.
We call this operation of moving the MGD to the next boundary the {\it leap} operation.
DX-NES-ICI performs the leap operation so that the new marginal probability is equal to \(\alpha\), as shown in \figref{fig:mywayAlgorithm:update1}. When \([\textbf{m}]_j \le \ell_{j,1|2}\) or \(\ell_{j,K_j-1|K_j}<[\textbf{m}]_j \), DX-NES-ICI performs the mean correction method given by \eqaref{eq:cmaeswm:modifyM} as CMA-ES w. Margin does.
The detail of the leap operation is explained in \secref{sec:proposal:leap}.

\begin{figure}[tb]
    \centering
    \begin{tabular}{cc}
        \begin{minipage}[t]{0.46\hsize}
            \centering
            \includegraphics[width=39mm]{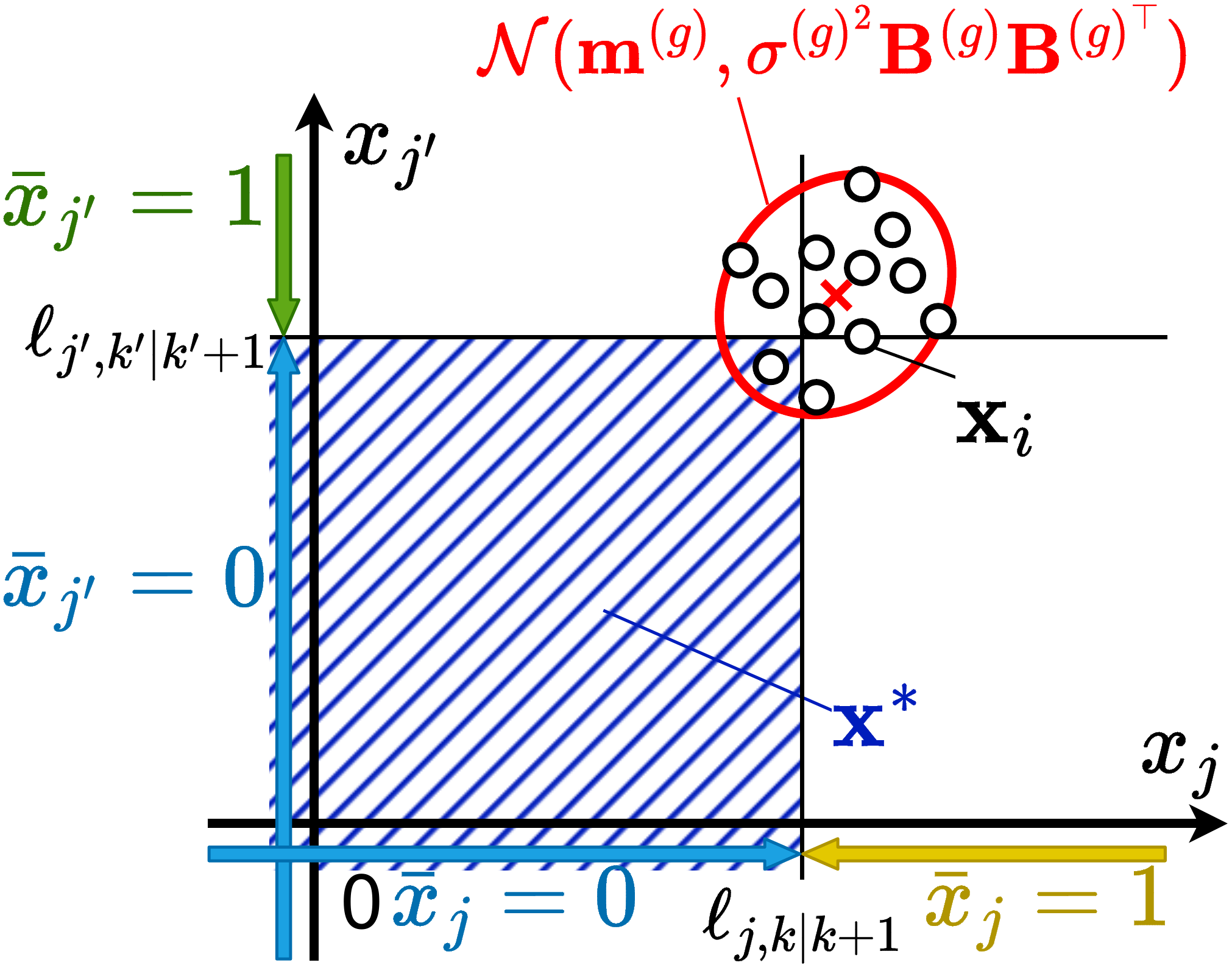}
            \subcaption{Many solutions are generated in not optimal regions according to MGD in $g$-th generation, and they are evaluated.}
            \label{fig:oneSideUnstable:1}
        \end{minipage}
         &
        \begin{minipage}[t]{0.46\hsize}
            \centering
            \includegraphics[width=39mm]{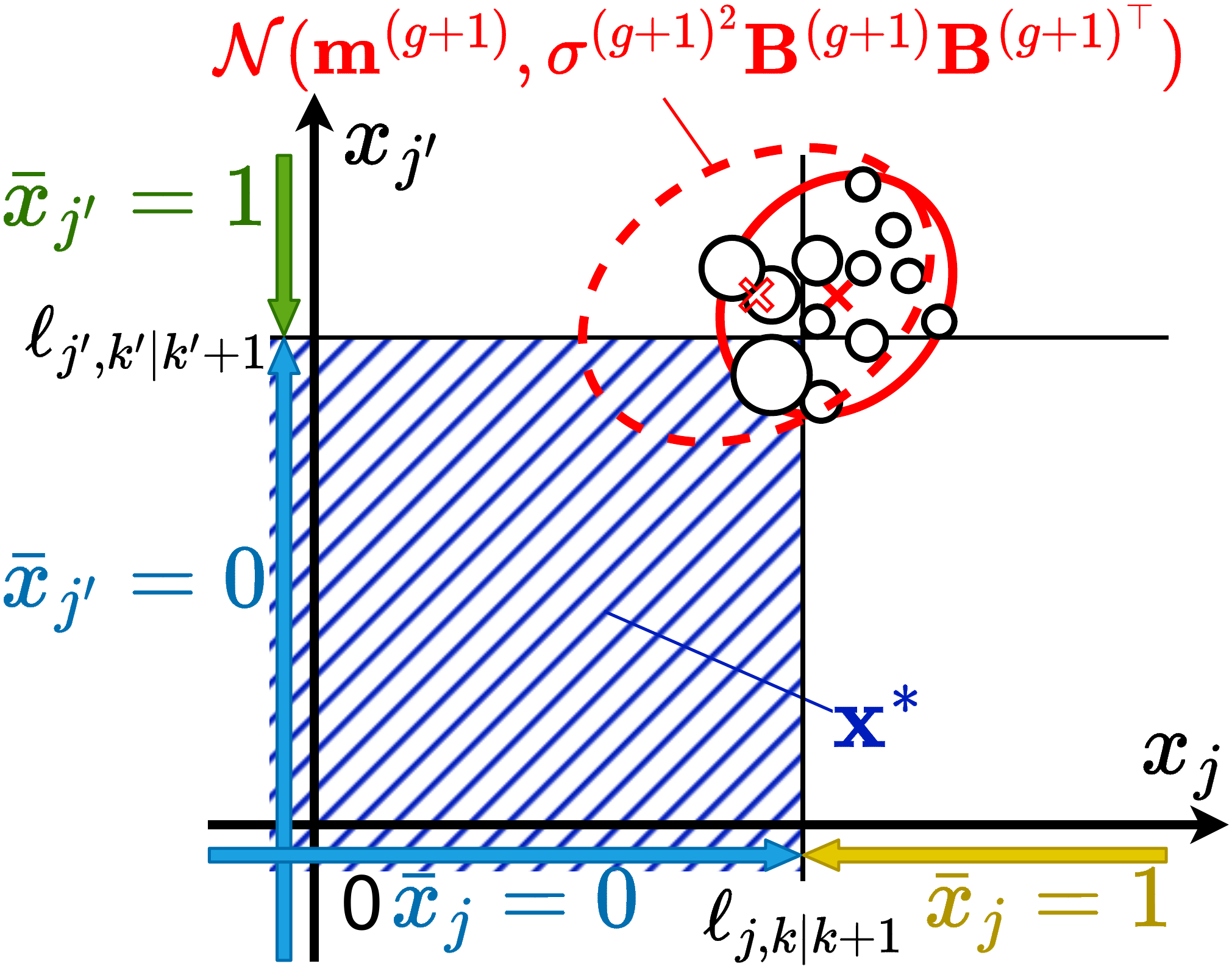}
            \subcaption{The size of the MGD of the next generation is expected to increase because the parameters of MGD are updated based on the better-evaluated solutions that exist on the outer extension part of MGD.}
            \label{fig:oneSideUnstable:2}
        \end{minipage}
    \end{tabular}
    \caption{
        Behavior of DX-NES-IC+Leap, when applied to an integer-ill-scaled function.
    	The vertical axis, the horizontal axis, and \(\textbf{x}^*\) represent the dimension of an integer variable \(\bar{x}_{j'}\), the dimension of another integer variable \(\bar{x}_{j}\), and the optimal region, respectively, where \(\bar{x}_j\) contributes more to the objective function value than \(\bar{x}_{j'}\).
    }
    \label{fig:oneSideUnstable}
\end{figure}

However, a method, which is DX-NES-IC incorporated with the transformation function \(\text{Encoding}_f\), the leap operation, and the mean correction, shows unstable behavior near the optimum as shown in \figref{fig:oneSideUnstable} when applied to functions where the contribution of each integer variable to the objective function value is very different from each other.
We call the method {\it DX-NES-IC+Leap}, and such functions {\it integer-ill-scaled functions} for simplicity, respectively, in this paper.
As shown in \figref{fig:oneSideUnstable:1}, the mean vector of MGD may leave from the optimal plateau due to the effect of random numbers after it reaches the optimal plateau and the size of MGD becomes smaller than that of the plateau.
In this situation, many solutions are generated in non-optimal plateaus according to the MGD, and the solutions generated on the outer extension part of the MGD could be better than those near the mean vector.
As a result, the MGD will be updated to expand as shown in \figref{fig:oneSideUnstable:2}.
This phenomenon should be more likely to occur when the dimension of the objective function is higher.
Therefore, the expansion of the MGD could continue over several generations when DX-NES-IC+Leap is applied to high-dimensional integer-ill-scaled functions.

\begin{figure}[tb]
    \centering
    \includegraphics[width=39mm]{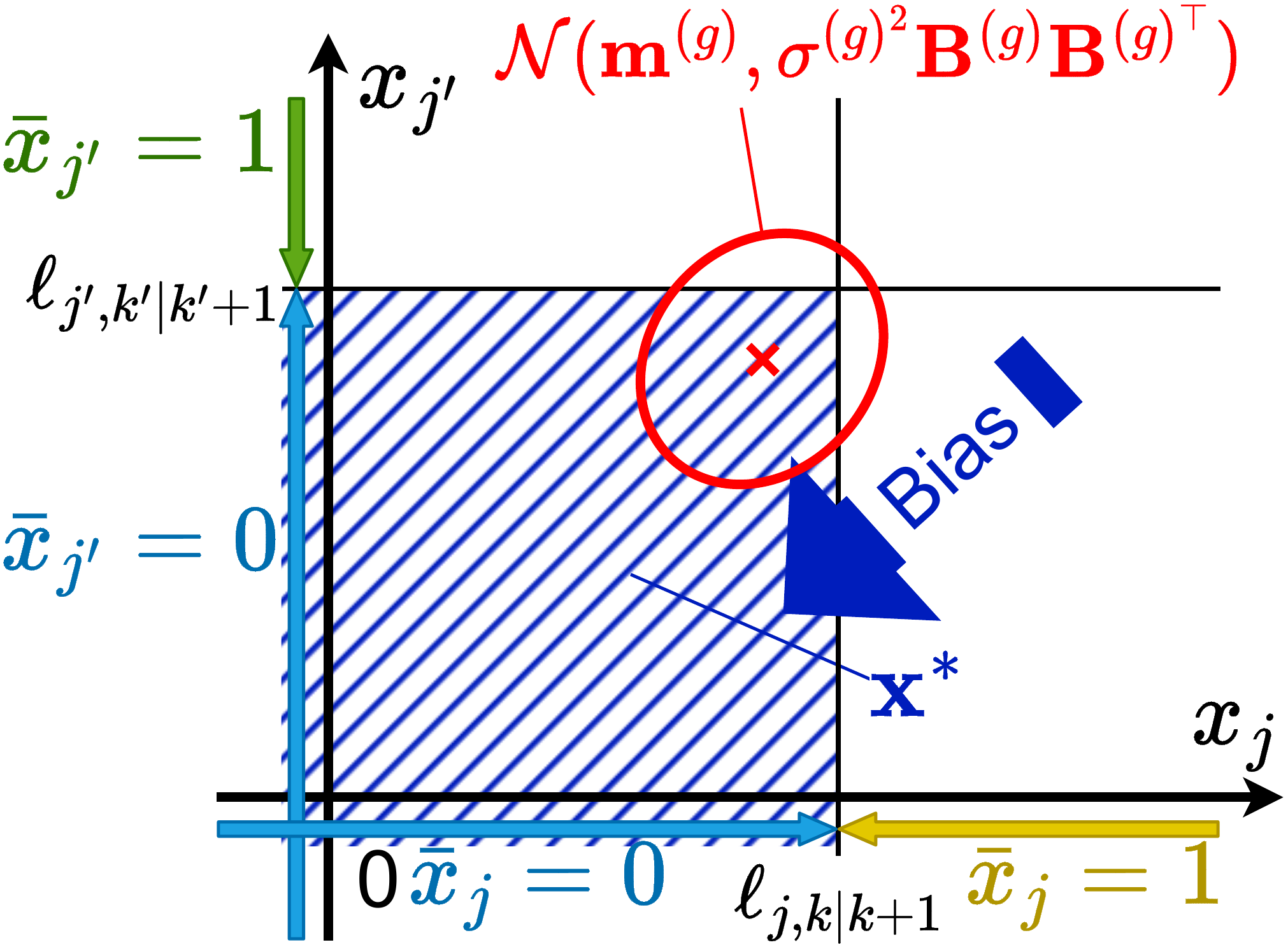}
    \caption{
    How DX-NES-ICI biases the movement of the mean vector for suppressing the expansion of the MGD near the optimal plateau.
    Since the bias is expected to keep MGD from deviating from the optimal region in the dimensions of integer variables where the MGD has already converged, the step-size tends not to increase.
    The blue arrow represents the bias.
    }
    \label{fig:mywayStable}
\end{figure}

In order to suppress the expansion of the MGD near the optimal plateau, we propose to bias the movement of the mean vector of the MGD towards the direction of a better plateau as shown in \figref{fig:mywayStable}.
Since the bias keeps MGD from deviating from the optimal plateau in the dimensions of integer variables where the MGD has already converged, the size of MGD is expected not to increase.
The detail of how to bias the movement of the mean vector is explained in \secref{sec:proposal:bias}.

\subsection{Leap operation of the mean vector}
\label{sec:proposal:leap}
The leap operation moves the mean vector of the MGD largely if the marginal probabilities of both sides of the MGD are less than \(\alpha\) in order to prevent the MGD from being trapped in a plateau.
The leap operation is defined by
\begin{eqnarray}
    [\textbf{m}^\text{after}]_j\leftarrow
    \left\{
    \begin{array}{ll}
        \ell_\text{low} \left([\textbf{m}^{\text{after}}]_j\right)+\textrm{CI}_j(\boldsymbol{\Sigma}^\text{after},\alpha),                          \\
        \;\;\;\;\;\;\;\;\;\;\; \textrm{ if } [\textbf{m}^{\text{after}}]_j\le \ell_\text{close} \left([\textbf{m}^{\text{before}}]_j\right), \\
        \ell_\text{up} \left([\textbf{m}^{\text{after}}]_j\right)-\textrm{CI}_j(\boldsymbol{\Sigma}^\text{after},\alpha),                           \\
        \textrm{
            \;\;\;\;\;\;\;\;\;\;\;Otherwise. }                                                                                                \\
    \end{array}
    \right.
\end{eqnarray}
where \(\textbf{m}^{\text{before}}\) is the mean vector of the MGD before updated by DX-NES-IC, and \(\textbf{m}^{\text{after}}\) and \(\boldsymbol{\Sigma}^\text{after}\) are the mean vector and the covariance matrix of the MGD after updated by DX-NES-IC, respectively.
It is easy to determine whether the marginal probabilities of both sides of the MGD are less than \(\alpha\) or not by using the resolution \({\rm Res}_j (\textbf{m}, \boldsymbol{\Sigma})\).
The resolution \({\rm Res}_j (\textbf{m}, \boldsymbol{\Sigma})\) is
the number of thresholds within \(\left[[\textbf{m}]_j - \textrm{CI}_j(\boldsymbol{\Sigma}, \alpha), [\textbf{m}]_j + \textrm{CI}_j(\boldsymbol{\Sigma}, \alpha)\right]\).
If \({\rm Res}_j (\textbf{m}, \boldsymbol{\Sigma}) = 0\), the marginal probabilities of both sides of the MGD are less than \(\alpha\).

Note that, when there is the marginal probability on a single side, the mean vector is modified by the mean correction method defined by \eqaref{eq:cmaeswm:modifyM} as CMA-ES w. Margin does.

\subsection{Bias of the movement of the mean vector}
\label{sec:proposal:bias}
DX-NES-ICI tries to prevent MGD from deviating from the optimal plateau by increasing the amount of movement when the mean vector moves in the direction of the center of the optimal plateau in the dimension of an integer variable where MGD has already converged to the optimal plateau.
When the MGD converges to the optimal plateau in the \(j\)-th dimension, MGD is located on only one threshold and thus satisfies that the resolution is less than one.
In the dimension where MGD is converging to the optimal plateau, the direction of the center of the optimal plateau to which the mean vector should move is away from the nearest threshold, as shown in \figref{fig:mywayStable}.
Since this orientation can be computed by the sign of \([\textbf{m}^{(g)}]_j-\ell_\text{close}([\textbf{m}^{(g)}]_j)\), DX-NES-ICI increases the learning rate of the average vector when this sign matches the sign of the update amount of the average vector.
In this paper, the learning rate of the mean vector is a vector \(\bm{\eta}_\textbf{m}\) because we want to manipulate the movement amount of the mean vector in each dimension.
Accordingly, the update rule for the mean vector is redefined as follows.
\begin{eqnarray}
    \textbf{m}^{(g+1)}\leftarrow \textbf{m}^{(g)}+\sigma^{(g)}\bm{\eta}_\textbf{m}^{(g)}\odot\textbf{B}^{(g)}\textbf{G}_\delta,
\end{eqnarray}
where the initial value of the learning rate of the mean vector is one, \(\bm{\eta}_\textbf{m}=\textbf{1}\).

\subsection{Algorithm}
\label{sec:proposal:algorithm}
\algoref{alg:proposal} shows the algorithm of the proposed method, DX-NES-ICI.
Our code is available at \url{https://github.com/ono-lab/dxnesici}.
The recommended values of the hyperparameters of DX-NES-ICI are shown in \tabref{tbl:dxnesic:defaultParameter}.
\begin{table}[tb]
    \centering
    \caption{The recommended values of the hyperparameters of DX-NES-ICI}
    \label{tbl:dxnesic:defaultParameter}
    \begin{tabular}{c|c} \hline
        \(w_i^\text{rank}\) & \(\frac{\hat{w}_i}{\sum_{j=1}^\lambda \hat{w}_j}-\frac{1}{\lambda}\)                         \\
        \(\hat{w}_i\)       & \(\max\left(0,\ln\left(\frac{\lambda}{2}+1\right)-\ln i\right)\)                             \\
        \(\mu_\text{eff}\)  & \(1/\sum_{i=1}^\lambda\left(w_i^\text{rank}+\frac{1}{\lambda}\right)^2\)                     \\
        \(c_\sigma\)        & \(\frac{1}{2\ln(N+1)}\frac{\mu_\textrm{eff}+2}{N+\mu_\textrm{eff}+5}\)                       \\
        \(\eta_\textbf{m}\) & \(\textbf{1}\)                                                                               \\
        \(\epsilon\)        & \(\mathbb{E}[||\mathcal{N}(\textbf{0},\textbf{I})||]\approx \sqrt{N}\{1-1/(4N)+1/(21N^2)\}\) \\\hline
    \end{tabular}
\end{table}
In \algoref{alg:proposal},
the dimension of the objective function \(N\), the objective function \(f\), the initial mean vector \(\textbf{m}^{(0)}\), the initial standard deviation \(\sigma^{(0)}\), the population size \(\lambda\), the minimum marginal probability \(\alpha\) are given as inputs, where \(\alpha=1/(N\cdot\lambda)\) \cite{cmaeswm}.
In line 1, the normalized transformation matrix \(\textbf{B}^{(0)}\), the evolution path \(\textbf{p}_\sigma^{(0)}\), and the generation \(g\) are initialized as \(\textbf{I}\in\mathbb{R}^{N\times N}\), \(\textbf{0}\), and \(0\), respectively.
Lines 3-37 are the operations for generation \(g\).
In lines 3-7, the population is generated according to the MGD \(\mathcal{N}(\textbf{m}^{(g)}, \sigma^{(g)^2}\textbf{B}^{(g)}\textbf{B}^{(g)^\top})\).
In line 8, the population \(\{\bar{\textbf{x}}_i\}_{i=1}^\lambda\) is evaluated, and it is sorted based on the objective function value \(f(\bar{\textbf{x}}_i)\).
In line 9, the evolution path \(\textbf{p}_\sigma^{(0)}\) is updated.
In line 10, calculateWeights determines the weight function \(w_i^{(g)}\) at generation \(g\) based on the norm of the evolution path.
In line 11, calculateLearningRate determines the learning rate of the standard deviation \(\eta_\sigma^{(g)}\) and that of the normalized transformation matrix \(\eta_\textbf{B}^{(g)}\) according to the norm of the evolution path.
In lines 12 and 13, the gradients of the parameters of the MGD \(\sigma^{(g)}\), \(\textbf{B}^{(g)}\), \(\textbf{m}^{(g)}\) are calculated by using the population \(\{\bar{\textbf{x}}_i\}_{i=1}^\lambda\) weighted according to the order \(i\) of the objective function value \(f(\bar{\textbf{x}}_i)\).
In lines 14-19, the bias operation is applied to the movement of the mean vector as described in \secref{sec:proposal:bias}.
In lines 20-22, the parameters of the MGD \(\sigma^{(g)}\), \(\textbf{B}^{(g)}\), \(\textbf{m}^{(g)}\) are updated.
In lines 23-25, the magnification of the MGD is emphasized by emphasizeExpansion if the norm of the evolution path is larger than \(\epsilon\), where emphasizeExpansion takes the normalized transformation matrix at the current generation \(\textbf{B}^{(g)}\), the normalized transformation matrix at the next generation \(\textbf{B}^{(g+1)}\), and the standard deviation at the next generation \(\sigma^{(g+1)}\) as the arguments, and returns the modified normalized transformation matrix at the next generation \(\textbf{B}^{(g+1)}\) and the modified standard deviation at the next generation \(\sigma^{(g+1)}\).
In lines 26-36, the leap operation and the mean correction are performed as described in \secref{sec:proposal:leap}.
See \cite{dxnesic} for the detailed algorithms of calculateWeights, calculateLearningRate, and emphasizeExpansion.
\begin{algorithm}[tb]
    \caption{Algorithm of DX-NES-ICI}
    \label{alg:proposal}
    {
        \fontsize{6.9}{6.9}\selectfont
        \begin{algorithmic}[1]
            \REQUIRE \(N\in\mathbb{N}\), \(f\), \(\textbf{m}^{(0)}\in\mathbb{R}^N\), \(\sigma^{(0)}\in\mathbb{R}\), \(\lambda\in\{2n|n\in\mathbb{N}\}\), \(\alpha\leftarrow 1/(N\cdot\lambda)\)
            \STATE  \(\textbf{B}^{(0)}\leftarrow\textbf{I}\in\mathbb{R}^{N\times N}\), \(\textbf{p}_\sigma^{(0)}\leftarrow \textbf{0}\), \(g\leftarrow 0\)
            \WHILE {{\rm termination condition not met}}
            \FOR {\(i \in \{1,\dots,\lambda/2\} \)}
            \STATE \(\textbf{z}_{2i-1}\leftarrow\mathcal{N}(\textbf{0},\textbf{I})\), \(\textbf{z}_{2i}\leftarrow -\textbf{z}_{2i-1}\)
            \STATE \(\textbf{x}_{2i-1}\leftarrow\textbf{m}^{(g)}+\sigma^{(g)}\textbf{B}^{(g)}\textbf{z}_{2i-1}\), \(\textbf{x}_{2i}\leftarrow\textbf{m}^{(g)}+\sigma^{(g)}\textbf{B}^{(g)}\textbf{z}_{2i}\)
            \STATE \(\bar{\textbf{x}}_{2i-1}\leftarrow {\rm Encoding}_f (\textbf{x}_{2i-1}),\bar{\textbf{x}}_{2i}\leftarrow {\rm Encoding}_f (\textbf{x}_{2i})\)
            \ENDFOR
            \STATE {\rm sort \(\{(\textbf{z}_i,\textbf{x}_i)\}_{i=1}^\lambda\) with respect to \(f(\bar{\textbf{x}}_i)\)}
            \STATE \(\textbf{p}_\sigma^{(g+1)}\leftarrow (1-c_\sigma)\textbf{p}_\sigma^{(g)}+\sqrt{c_\sigma(2-c_\sigma)\mu_{\rm eff}}\sum_{i=1}^\lambda w_i^{\rm rank}\textbf{z}_i\)
            \STATE \(w_i^{(g)}\leftarrow \text{calculateWeights}(||\textbf{p}_\sigma^{(g+1)}||)\)
            \STATE \(\eta_\sigma^{(g)},\eta_\textbf{B}^{(g)}\leftarrow \text{calculateLearningRate}(||\textbf{p}_\sigma^{(g+1)}||)\)
            \STATE \(\textbf{G}_\textbf{M}\leftarrow \sum_{i=1}^\lambda w_i^{(g)}(\textbf{z}_i\textbf{z}_i^\top-\textbf{I})\), \(G_\sigma\leftarrow {\rm tr}(\textbf{G}_\textbf{M}/N)\), \(\textbf{G}_\textbf{B}\leftarrow \textbf{G}_\textbf{M} - G_\sigma\textbf{I}\)
            \STATE \(\textbf{G}_\delta\leftarrow\sum_{i=1}^\lambda w_i^{(g)}\textbf{z}_i\)
            \STATE \(\bm{\eta}_\textbf{m}^{(g)}\leftarrow \textbf{1}\)
            \FOR {\(j=N_{\rm co}+1,\ldots,N\)}
            \IF {${\rm Res}^{(g)}(j) \le 1 \land $ \\
                \;\;\;\;${\rm sign}\left([\textbf{G}_\delta]_j\right) = {\rm sign}\left([\textbf{m}^{(g)}]_j - \ell_{\rm close}([\textbf{m}^{(g)}]_j)\right)$}
            \STATE \([\bm{\eta}_\textbf{m}^{(g)}]_j \leftarrow [\bm{\eta}_\textbf{m}^{(g)}]_j + 1\)
            \ENDIF
            \ENDFOR
            \STATE \(\sigma^{(g+1)}\leftarrow\sigma^{(g)}{\rm exp}\left(\eta_\sigma^{(g)}G_\sigma/2\right)\)
            \STATE \(\textbf{B}^{(g+1)}\leftarrow\textbf{B}^{(g)}{\rm exp}\left(\eta_\textbf{B}^{(g)}\textbf{G}_\textbf{B}/2\right)\)
            \STATE \(\textbf{m}^{(g+1)}\leftarrow \textbf{m}^{(g)}+\sigma^{(g)}\bm{\eta}_\textbf{m}^{(g)}\odot\textbf{B}^{(g)}\textbf{G}_\delta\)
            \IF {\(||\textbf{p}_\sigma^{(g+1)}||\ge\epsilon\)}
            \STATE \(\sigma^{(g+1)},\textbf{B}^{(g+1)}\leftarrow \text{emphasizeExpansion}(\textbf{B}^{(g)}, \sigma^{(g+1)}, \textbf{B}^{(g+1)})\)
            \ENDIF
            \FOR {\(j=N_{\rm co}+1,\ldots,N\)}
            \IF {\({\rm Res}^{(g+1)}(j) = 0\)}
            \IF {\([\textbf{m}^{(g+1)}]_j\le\ell_{j,1|2}\lor\ell_{j,K_j-1|K_j} < [\textbf{m}^{(g+1)}]_j\)}
            \STATE $[\textbf{m}^{(g+1)}]_j\leftarrow \ell_{\rm close} \left([\textbf{m}^{(g+1)}]_j\right)+$\\
                $\;\;\;\;\;\;\;\;\;\;\;\;\;\;\;\;\;\;\;\;\;{\rm sign}\left([\textbf{m}^{(g+1)}]_j-\ell_{\rm close}\left([\textbf{m}^{(g+1)}]_j\right)\right)$\\
                $\;\;\;\;\;\;\;\;\;\;\;\;\;\;\;\;\;\;\;\;\;\times{\rm CI}_j\left(\sigma^{(g+1)^2}\textbf{B}^{(g+1)}\textbf{B}^{(g+1)^\top}, \alpha\right)$
            \ELSIF {\([\textbf{m}^{(g+1)}]_j\le \ell_{\rm close} \left([\textbf{m}^{(g)}]_j\right)\)}
            \STATE \([\textbf{m}^{(g+1)}]_j\leftarrow \ell_{\rm low} \left([\textbf{m}^{(g+1)}]_j\right)+{\rm CI}_j\left(\sigma^{(g+1)^2}\textbf{B}^{(g+1)}\textbf{B}^{(g+1)^\top}, \alpha\right)\)
            \ELSE
            \STATE \([\textbf{m}^{(g+1)}]_j\leftarrow \ell_{\rm up} \left([\textbf{m}^{(g+1)}]_j\right)-{\rm CI}_j\left(\sigma^{(g+1)^2}\textbf{B}^{(g+1)}\textbf{B}^{(g+1)^\top}, \alpha\right)\)
            \ENDIF
            \ENDIF
            \ENDFOR
            \STATE \(g\leftarrow g+ 1\)
            \ENDWHILE
        \end{algorithmic}
    }
\end{algorithm}

\section{Experiment}\label{sec:exp}
\subsection{Purpose}
We compare the search performance of DX-NES-ICI with that of CMA-ES w. Margin using benchmark problems where continuous variables contribute more to the objective function value than integer ones and the other benchmark problems where CMA-ES w. Margin originally showed good performance.

\subsection{Benchmark problems}
\label{sec:exp:benchmark}
In this experiment, we use the following benchmark problems for the dimension \(N=20,40,80\).
\begin{itemize}
  \item \(N_\text{int}\)-tablet(\(\bar{\textbf{x}}\))\(=\sum_{j=1}^{N_\text{int}} [\bar{\textbf{x}}_\text{int}]_j^2 + \sum_{j=1}^{N_\text{co}} \left(100 [\bar{\textbf{x}}_\text{co}]_j\right)^2\)
  \item \(\text{ReversedEllipsoidInt}(\bar{\textbf{x}})=\)\\
        \(\sum_{j=1}^{N_\text{int}} (1000^\frac{j-1}{N-1}[\bar{\textbf{x}}_\text{int}]_j)^2 + \sum_{j=1}^{N_\text{co}} (1000^\frac{N_\text{int}+j-1}{N-1}[\bar{\textbf{x}}_\text{co}]_j)^2\)
  \item EllipsoidInt(\(\bar{\textbf{x}}\))\(=\sum_{j=1}^{N} (1000^\frac{j-1}{N-1}[\bar{\textbf{x}}]_j)^2\)
  \item SphereOneMax(\(\bar{\textbf{x}}\)) \(=\sum_{j=1}^{N_\text{co}} ([\bar{\textbf{x}}_\text{co}]_j)^2 + N_\text{int}-\sum_{j=1}^{N_\text{int}} [\bar{\textbf{x}}_\text{int}]_j\)
\end{itemize}
\(N_\text{int}\)-tablet and ReversedEllipsoidInt are MI-BBO benchmark functions where continuous variables contribute more to the objective function value than integer ones.
EllipsoidInt is an MI-BBO benchmark function where integer variables contribute more to the objective function value than continuous ones.
SphereOneMax is a 0-1 MI-BBO benchmark function.
CMA-ES w. Margin has been reported to show good performance for EllipsoidInt and SphereOneMax\cite{cmaeswm}.
In all the MI-BBO benchmark functions, the domain of an integer variable is given by \(\{-10,-9,\ldots,9,10\}\), which is also used in \cite{cmaeswm}, and the optimal solution is \(\bar{\textbf{x}}^*=\textbf{0}\).
In the 0-1 MI-BBO benchmark function, the domain of an integer variable is given by \(\{0,1\}\), and the optimal solution is \(\bar{\textbf{x}}_\text{co}^*=\textbf{0}\land \bar{\textbf{x}}_\text{int}^*=\textbf{1}\).
\begin{table*}[tb]
  \centering
  \caption{Results of DX-NES-ICI and CMA-ES w. Margin on the benchmark functions for \(N=20,40,80\).
    DX-NES-ICI is the proposed method.
    "\#Suc." represents the number of successful trials out of 100 trials. "\#Eval." represents the average number of evaluations in successful trials.
    The ratio of \#Eval. (DX-NES-ICI / CMA-ES w. Margin) is shown when each "\#Suc." is 100 in both methods.
    The interquartile range (IQR) and \(\lambda\) are listed alongside the result obtained with those.
    The bold text indicates the best \#Suc. and \#Eval.}
  \label{tab:exp:mixedInt}
  \begin{tabular}{c|c|cc|cc|c|}
                            &       & \multicolumn{2}{c|}{DX-NES-ICI} & \multicolumn{2}{c|}{CMA-ES w. Margin} &                                                                   \\
    Function                & \(N\) & \#Suc.                        & \#Eval. (IQR) (\(\lambda\))           & \#Suc.           & \#Eval. (IQR) (\(\lambda\)) & \#Eval. Ratio   \\\hline\hline
                            & 20    & \textbf{100/100}              & \textbf{3111} (286) (6)               & 80/100           & 19078 (868) (28)            & -                \\
    \(N_\text{int}\)-tablet & 40    & \textbf{100/100}              & \textbf{6356} (540) (8)               & 42/100           & 31574 (2272) (30)           & -                \\
                            & 80    & \textbf{100/100}              & \textbf{13305} (810) (12)             & 28/100           & 57950 (6556) (22)           & -                \\\hline
                            & 20    & \textbf{100/100}              & \textbf{5202} (477) (10)              & 58/100           & 17237 (1920) (30)           & -                \\
    ReversedEllipsoidInt    & 40    & \textbf{100/100}              & \textbf{12714} (1333) (14)            & 33/100           & 43975 (4530) (30)           & -                \\
                            & 80    & \textbf{100/100}              & \textbf{34340} (2434) (18)            & 27/100           & 108425 (27255) (10)         & -                \\\hline
                            & 20    & \textbf{100/100}              & \textbf{6306} (963) (12)              & \textbf{100/100} & 8123 (1038) (10)            & \(\times 0.78 \) \\
    EllipsoidInt            & 40    & \textbf{100/100}              & \textbf{15758} (1884) (16)            & \textbf{100/100} & 22923 (2400) (16)           & \(\times 0.69 \) \\
                            & 80    & \textbf{100/100}              & \textbf{42135} (4163) (22)            & \textbf{100/100} & 66851 (5010) (20)           & \(\times 0.63 \) \\\hline
                            & 20    & \textbf{100/100}              & \textbf{1962} (244) (8)               & \textbf{100/100} & 3757 (350) (10)             & \(\times 0.52 \) \\
    SphereOneMax            & 40    & \textbf{100/100}              & \textbf{3878} (280) (10)              & \textbf{100/100} & 7975 (536) (14)             & \(\times 0.49 \) \\
                            & 80    & \textbf{100/100}              & \textbf{8133} (406) (14)              & \textbf{100/100} & 17145 (902) (22)            & \(\times 0.47 \) \\\hline
  \end{tabular}
\end{table*}

\subsection{Performance measures}
\label{sec:exp:index}
We assess the performance of each method using two measures in order.
We use the maximum number of successful trials out of 100, changing population size \(\lambda\) as the first measure.
Each run is regarded as success if the solution whose evaluation value is smaller than \(10^{-10}\) is obtained before being regarded as failure.
Each run is regarded as failure when the number of evaluations exceeds \(N\times 10^4\), the minimum eigenvalue of a covariance matrix is less than \(10^{-30}\), or the condition number of the covariance matrix exceeds \(10^{14}\).
We use the minimum average number of evaluations in successful trials under the condition that the maximum number of successful trials is obtained.

\subsection{Experimental settings}
The candidates of \(\lambda\) are 6, 8, 10, 12, 14, 16, 18, 20, 22, 24, 26, 28, and 30, where the range of the recommended value \(4+\lfloor3\ln N\rfloor\) in CMA-ES w. Margin is included.
We use \(\sigma^{(0)}=1\) and \(\alpha=1/(N\cdot\lambda)\).
Each element of an initial mean vector \(\textbf{m}^{(0)}\) is set to 0.5 in the dimensions of integer variables of a 0-1 MI-BBO function and a uniform random value from \([1,3]\) in the other dimensions.
Except for changing \(\lambda\), these experimental settings are also used in \cite{cmaeswm}.
We use the hyperparameters of CMA-ES w. Margin as default values set in \cite{cmaeswm} and those of DX-NES-ICI as the recommended values shown in \tabref{tbl:dxnesic:defaultParameter}.

\subsection{Results}
\tabref{tab:exp:mixedInt} shows the maximum number of successful trials and the minimum average number of evaluations in successful trials of DX-NES-ICI and CMA-ES w. Margin on the benchmark problems for \(N=20,40,80\).
The ratio of the minimum average number of evaluations (DX-NES-ICI / CMA-ES w. Margin) is shown when the maximum number of successful trials is 100 in both methods.
The smaller this ratio is, DX-NES-ICI is better than CMA-ES w. Margin.
The bold text indicates the best number of successful trials and the best average number of evaluations.
DX-NES-ICI improved from about 1.2 to 3.7 times the number of successful trials compared to CMA-ES w. Margin on benchmark problems where continuous variables contribute more to the objective function value than integer ones.
DX-NES-ICI also improved from about 0.47 to 0.78 times the average number of evaluations compared to CMA-ES w. Margin on benchmark problems where CMA-ES w. Margin originally showed good performance.

\section{Discussions}\label{sec:discussions}
\subsection{Effectiveness of the mechanism that leaps and corrects the mean vector}\label{sec:cons:1}
To confirm the effectiveness of the mechanism that leaps and corrects the mean vector, we compare the search performance of DX-NES-IC and that of DX-NES-IC+Leap.
The benchmark problems and the experimental settings are the same as in \secref{sec:exp}.
We show the results of the experiment in \tabref{tab:cons:1}.
\tabref{tab:cons:1} suggests that the mechanism is effective.
\begin{table}[tb]
    \centering
    \caption{Results of the DX-NES-IC+Leap and DX-NES-IC on 40-dimensional \(N_\text{int}\)-tablet and SphereOneMax, where the DX-NES-IC+Leap is the DX-NES-IC incorporated with the mechanism that leaps and corrects the mean vector.}
    \label{tab:cons:1}
    \begin{tabular}{c|c|c|c|}
                                &       & DX-NES-IC+Leap & DX-NES-IC \\
        Function                & \(N\) & \#Suc.     & \#Suc.    \\\hline\hline
        \(N_\text{int}\)-tablet & 40    & 100/100    & 0/100     \\\hline
        SphereOneMax            & 40    & 100/100    & 25/100    \\\hline
    \end{tabular}
\end{table}

\begin{figure}[tb]
	\centering
	\begin{tabular}{cc}
		\begin{minipage}[t]{0.46\hsize}
			\centering
			\includegraphics[width=39mm]{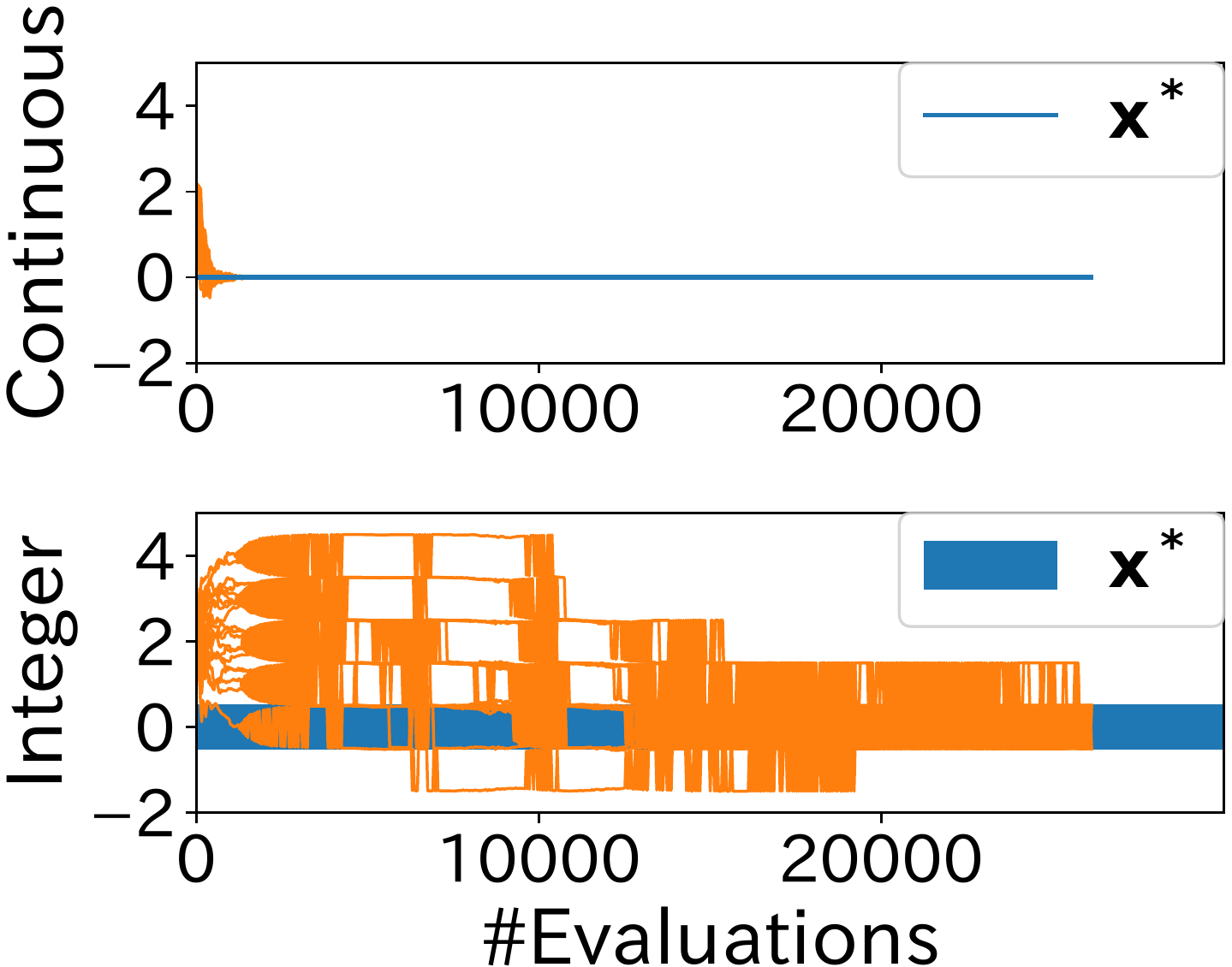}
			\subcaption{DX-NES-IC+Leap}
			\label{fig:NintTablet40:dxnesicLeap}
		\end{minipage}
		 &
		\begin{minipage}[t]{0.46\hsize}
			\centering
			\includegraphics[width=39mm]{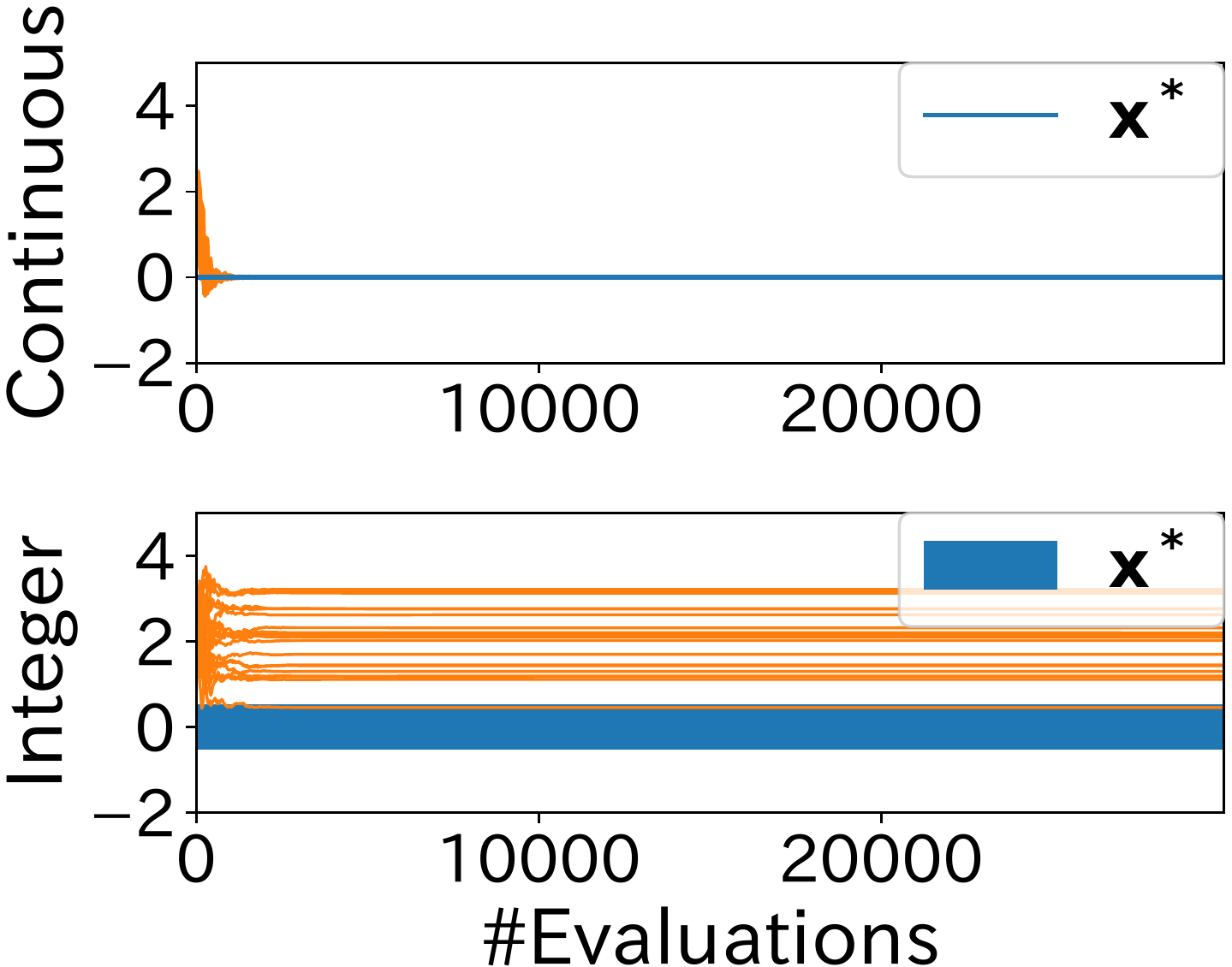}
			\subcaption{DX-NES-IC}
			\label{fig:NintTablet40:dxnesic}
		\end{minipage}
	\end{tabular}
	\caption{The transition plots of $[\textbf{m}]_j$ in the dimensions of continuous $(1\le j \le 20)$ and integer $(21\le j \le 40)$ variables on 40-dimensional $N_\text{int}$-tablet.
	The horizontal axis, the vertical axis, the orange curves, and the blue lines represent the number of evaluations, the value of $[\textbf{m}]_j$, the transition of $[\textbf{m}]_j$, and $\textbf{x}^*$, respectively.}
	\label{fig:NintTablet40}
\end{figure}

Figure \ref{fig:NintTablet40} shows the transition of $[\textbf{m}]_j$ of DX-NES-IC+Leap and DX-NES-IC on 40-dimensional $N_\text{int}$-tablet.
The experimental settings are the same as in \secref{sec:exp} except $\lambda=30$ and $\textbf{m}^{(0)}=2$.
As shown in \figref{fig:NintTablet40}, the mean vector moves largely in one generation in dimensions of integer variables in DX-NES-IC+Leap, which prevents it from stagnating, while it stagnates in DX-NES-IC.

\subsection{Effectiveness of the mechanism to bias the movement of the mean vector}
\label{sec:cons:2}
To confirm the effectiveness of the mechanism to bias the movement of the mean vector, we compare the search performance of DX-NES-ICI and that of DX-NES-IC+Leap, where the DX-NES-IC+Leap is DX-NES-ICI without the mechanism.
The benchmark problems and the experimental settings are the same as in \secref{sec:exp}.
We show the results of the experiment in \tabref{tab:cons:2}.
\tabref{tab:cons:2} suggests that the mechanism is effective.
\begin{table}[tb]
    \centering
    \caption{Results of DX-NES-ICI and DX-NES-IC+Leap on 80-dimensional ReversedEllipsoidInt and EllipsoidInt, where DX-NES-IC+Leap is DX-NES-ICI without the mechanism to bias the movement of the mean vector.}
    \label{tab:cons:2}
    \begin{tabular}{c|c|c|c|}
                             &       & DX-NES-ICI & DX-NES-IC+Leap \\
        Function             & \(N\) & \#Suc.   & \#Suc.     \\\hline\hline
        ReversedEllipsoidInt & 80    & 100/100  & 5/100      \\\hline
        EllipsoidInt         & 80    & 100/100  & 4/100      \\\hline
    \end{tabular}
\end{table}

\begin{figure}[tb]
    \centering
    \begin{tabular}{cc}
        \begin{minipage}[t]{0.46\hsize}
            \centering
            \includegraphics[width=39mm]{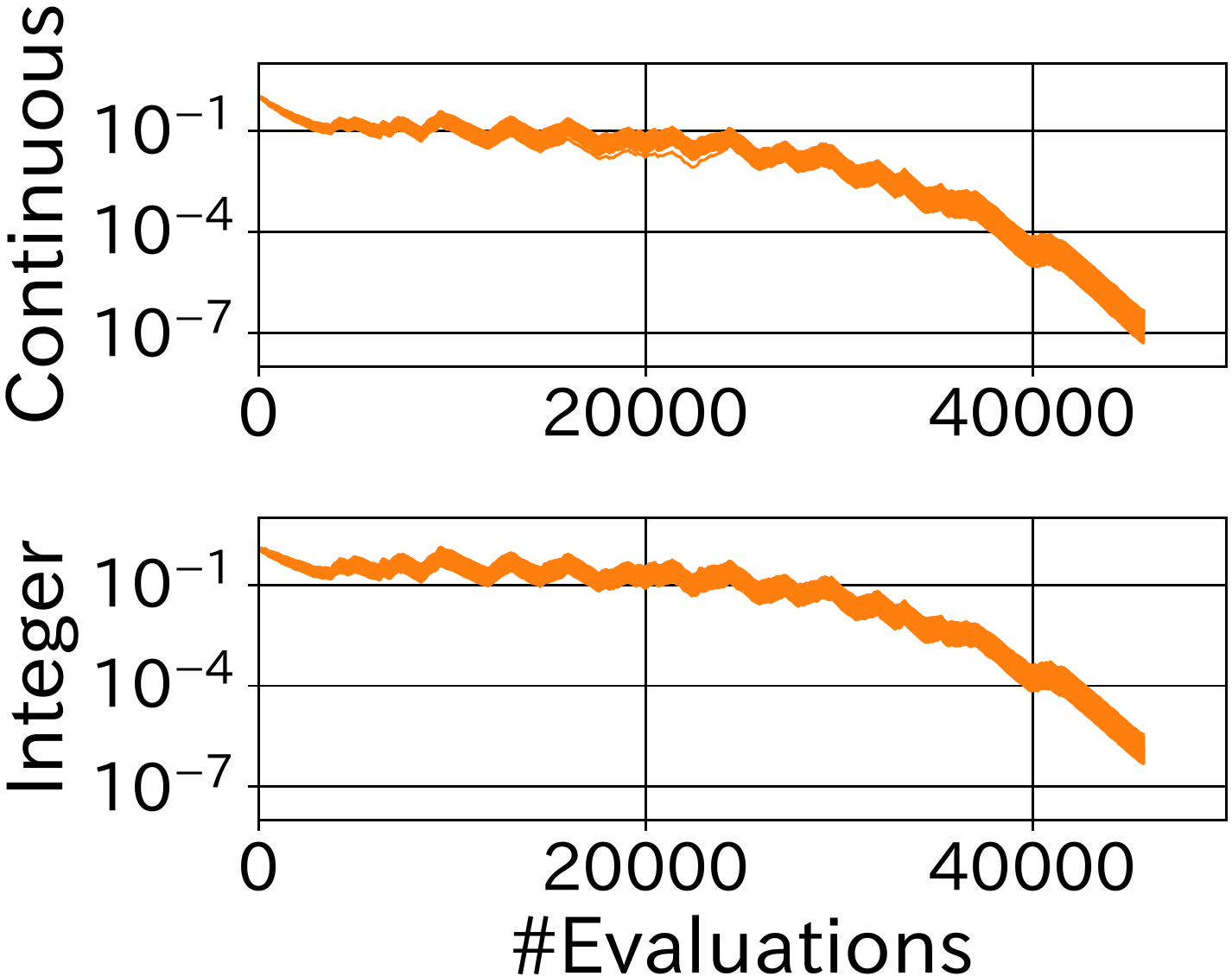}
            \subcaption{DX-NES-ICI}
            \label{fig:EllipsoidInt80:dxnesicProposed}
        \end{minipage}
         &
        \begin{minipage}[t]{0.46\hsize}
            \centering
            \includegraphics[width=39mm]{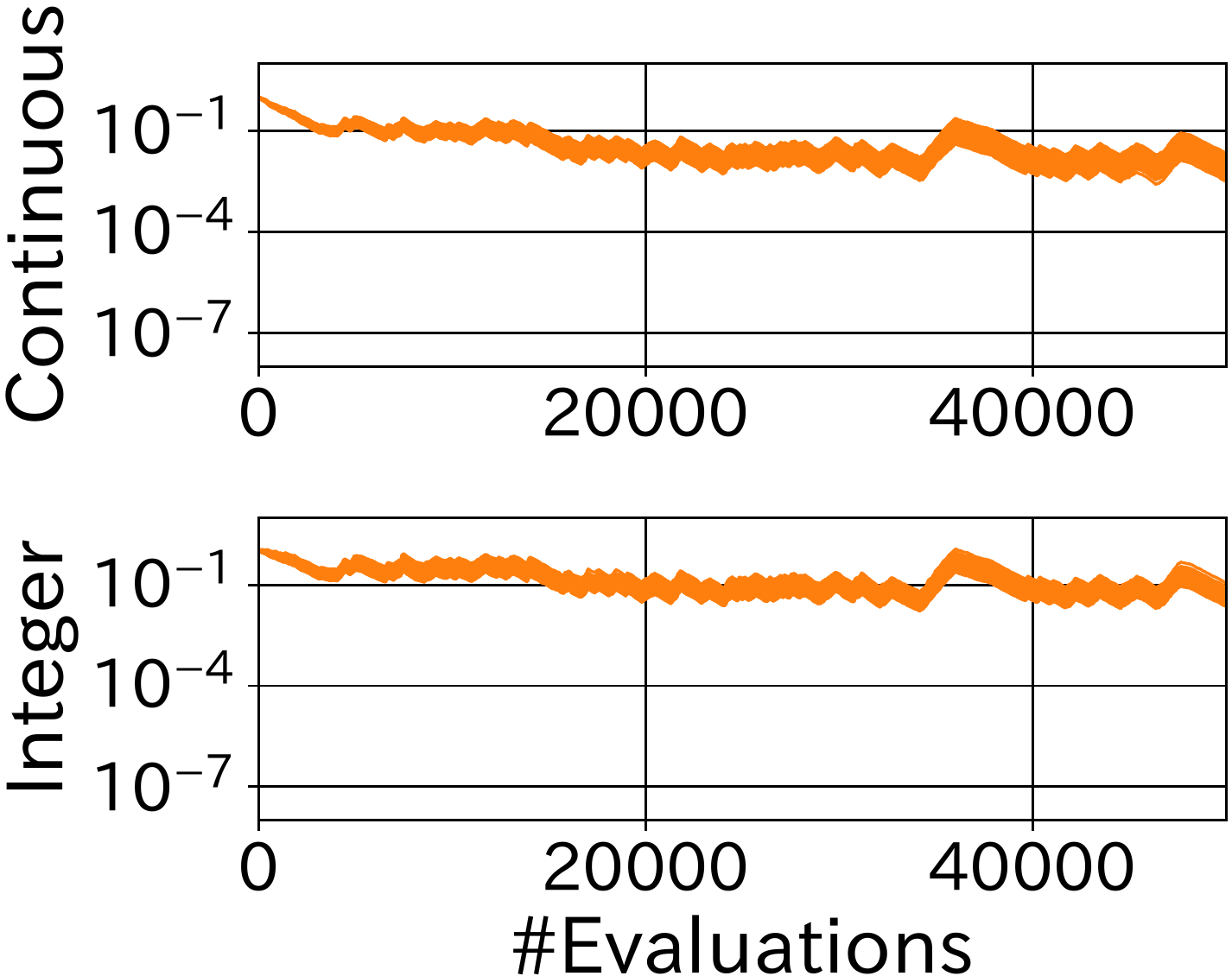}
            \subcaption{DX-NES-IC+Leap}
            \label{fig:EllipsoidInt80:dxnesicLeap}
        \end{minipage}
    \end{tabular}
    \caption{The transition plots of $\sigma\sqrt{\langle\textbf{B}\textbf{B}^\top\rangle_j}$ in the dimensions of continuous $(1\le j \le 40)$ and integer $(41\le j \le 80)$ variables on 80-dimensional EllipsoidInt.
	The horizontal axis, the vertical axis, and the orange curves represent the number of evaluations, the value of $\sigma\sqrt{\langle\textbf{B}\textbf{B}^\top\rangle_j}$, and the transition of $\sigma\sqrt{\langle\textbf{B}\textbf{B}^\top\rangle_j}$, respectively.}
    \label{fig:EllipsoidInt80}
\end{figure}

Figure \ref{fig:EllipsoidInt80} shows the transition of $\sigma\sqrt{\langle\textbf{B}\textbf{B}^\top\rangle_j}$ of DX-NES-ICI and DX-NES-IC+Leap on 80-dimensional EllipsoidInt.
The experimental settings are the same as in \secref{sec:exp} except $\lambda=30$ and $\textbf{m}^{(0)}=2$.
As shown in \figref{fig:EllipsoidInt80},
$\sigma\sqrt{\langle\textbf{B}\textbf{B}^\top\rangle_j}$ converges to zero in DX-NES-ICI, while it does not in DX-NES-IC+Leap.

\section{Conclusion}\label{sec:conclusion}
In this paper, we proposed a natural evolution strategy (NES) for mixed-integer black-box optimization (MI-BBO) that often appears in real-world problems such as hyperparameter optimization of machine learning and materials design.
This problem is difficult to optimize because plateaus where the values do not change appear when the integer variables are relaxed to the continuous ones.
CMA-ES w. Margin that addresses the plateaus reportedly showed good performance on MI-BBO benchmark problems.
However, it has been observed that the search performance of CMA-ES w. Margin deteriorates when continuous variables contribute more to the objective function value than integer ones.
In order to address the problem of CMA-ES w. Margin, we proposed Distance-weighted eXponential Natural Evolution Strategy taking account of Implicit Constraint and Integer (DX-NES-ICI), which is incorporated with the transformation function for real-value relaxation, the leap operation, the mean correction, and the bias operation of the movement of the mean vector.
We compared the search performance of DX-NES-ICI with that of CMA-ES w. Margin through numerical experiments.
As a result, DX-NES-ICI was up to 3.7 times better than CMA-ES w. Margin in terms of a rate of finding the optimal solutions on benchmark problems where continuous variables contribute more to the objective function value than integer ones.
DX-NES-ICI also outperformed CMA-ES w. Margin on problems where CMA-ES w. Margin originally showed good performance.
In future work, we would like to apply DX-NES-ICI to functions with epistasis among integer variables, evaluate the performance, and improve DX-NES-ICI if necessary.
We also have a plan to apply DX-NES-ICI to real-world applications such as lens design and show the effectiveness of DX-NES-ICI in real-world applications.
Moreover, we will confirm whether the proposed components are also effective for other ESs such as CMA-ES.

\end{document}